\pdfoutput=1

\documentclass[11pt]{article}

\usepackage[final]{acl}

\usepackage[dvipsnames]{xcolor}
\usepackage{times}
\usepackage{latexsym}
\usepackage{graphicx}
\usepackage{textcomp}
\usepackage{colortbl}
\usepackage{xcolor}
\usepackage{amsmath} 
\usepackage{booktabs} 
\usepackage{array}
\usepackage{graphicx} 
\usepackage{multirow}
\usepackage{color}
\usepackage{makecell} 
\usepackage{float}
\usepackage{subcaption}
\usepackage{colortbl}
\usepackage{placeins}
\usepackage[ruled,vlined]{algorithm2e}
\usepackage{comment}
\usepackage{pifont}
\usepackage{url} 
\usepackage{booktabs}
\usepackage{multirow}
\usepackage{tabularx} 
\usepackage{booktabs}
\usepackage{multirow}
\usepackage{adjustbox}
\usepackage{subcaption}
\usepackage[most]{tcolorbox}
\usepackage{stfloats}
\tcbuselibrary{listings} 
\tcbuselibrary{breakable} 
\usepackage{capt-of}
\usepackage{hyperref}

\usepackage{listings}
\newtcolorbox{promptbox}[1]{ 
  colback=gray!5,
  colframe=black!40,
  title=#1,
  breakable,
  sharp corners
}

\usepackage[T1]{fontenc}

\usepackage[utf8]{inputenc}

\usepackage{microtype}

\usepackage{inconsolata}

\usepackage{graphicx}

\usepackage{diagbox}
\usepackage{pgfplots}
\pgfplotsset{compat=1.18}
\usepackage{pgfplotstable}

\newcommand{\benchmarkname}{AgenticRAGTracer }
\title{AgenticRAGTracer: A Hop-Aware Benchmark for Diagnosing Multi-Step Retrieval Reasoning in Agentic RAG}

\author{
Qijie You$^{1}$\thanks{Equal contribution.}, 
Wenkai Yu$^{2}$\footnotemark[1],
{\bf Wentao Zhang$^{2,3,4}$} \\
$^{1}$University of Science and Technology Beijing \\
$^{2}$Peking University \\
$^{3}$Zhongguancun Academy \\
$^{4}$Beijing Key Laboratory of Data Intelligence and Security (Peking University) \\
\textit{u202342615@xs.ustb.edu.cn, \{ywk, wentao.zhang\}@pku.edu.cn} \\
}

\begin{document}
\maketitle
\begin{abstract}
With the rapid advancement of agent-based methods in recent years, Agentic RAG has undoubtedly become an important research direction. Multi-hop reasoning, which requires models to engage in deliberate thinking and multi-step interaction, serves as a critical testbed for assessing such capabilities. However, existing benchmarks typically provide only final questions and answers, while lacking the intermediate hop-level questions that gradually connect atomic questions to the final multi-hop query. This limitation prevents researchers from analyzing at which step an agent fails and restricts more fine-grained evaluation of model capabilities. Moreover, most current benchmarks are manually constructed, which is both time-consuming and labor-intensive, while also limiting scalability and generalization. To address these challenges, we introduce \benchmarkname, the first Agentic RAG benchmark that is primarily constructed automatically by large language models and designed to support step-by-step validation. Our benchmark spans multiple domains, contains 1,305 data points, and has no overlap with existing mainstream benchmarks. Extensive experiments demonstrate that even the best large language models perform poorly on our dataset. For instance, GPT-5 attains merely 22.6\% EM accuracy on the hardest portion of our dataset. Hop-aware diagnosis reveals that failures are primarily driven by distorted reasoning chains—either collapsing prematurely or wandering into over-extension. This highlights a critical inability to allocate steps consistent with the task's logical structure, providing a diagnostic dimension missing in traditional evaluations. We believe our work will facilitate research in Agentic RAG and inspire further meaningful progress in this area.
 Our code and data are available at \url{https://github.com/YqjMartin/AgenticRAGTracer}.
\end{abstract}

\section{Introduction}

With the rapid evolution of the agentic paradigm, Large Language Models (LLMs) have expanded beyond simple text generation to master complex workflows and broad tool usage. Among these capabilities, Retrieval-Augmented Generation (RAG) is pivotal, enabling LLMs to access information beyond their parametric memory, thereby enhancing generalization and mitigating hallucinations. Consequently, advancing the capabilities of Agentic RAG—systems that autonomously plan, retrieve, and reason—has become a central research focus.

Compared to single-hop queries (e.g., NQ~\citep{nq}, TriviaQA~\citep{triviaqa}), multi-hop questions serve as a rigorous testbed for Agentic RAG by demanding iterative reasoning and multi-document retrieval. However, current evaluations predominantly rely on benchmarks designed for traditional RAG, such as HotpotQA~\citep{hotpotqa}, 2WikiMultihopQA~\citep{2wiki}, and MuSiQue~\citep{musique}. These datasets generally provide only final QA pairs with short supporting passages, assuming a retrieve-then-read paradigm where all content is processed at once. For Agentic RAG, which inherently relies on dynamic, multi-step interactions, evaluating performance solely on the final outcome fails to capture the intricate decision-making process.
\begin{figure*}[h]
    \centering
    \includegraphics[width=1\linewidth]{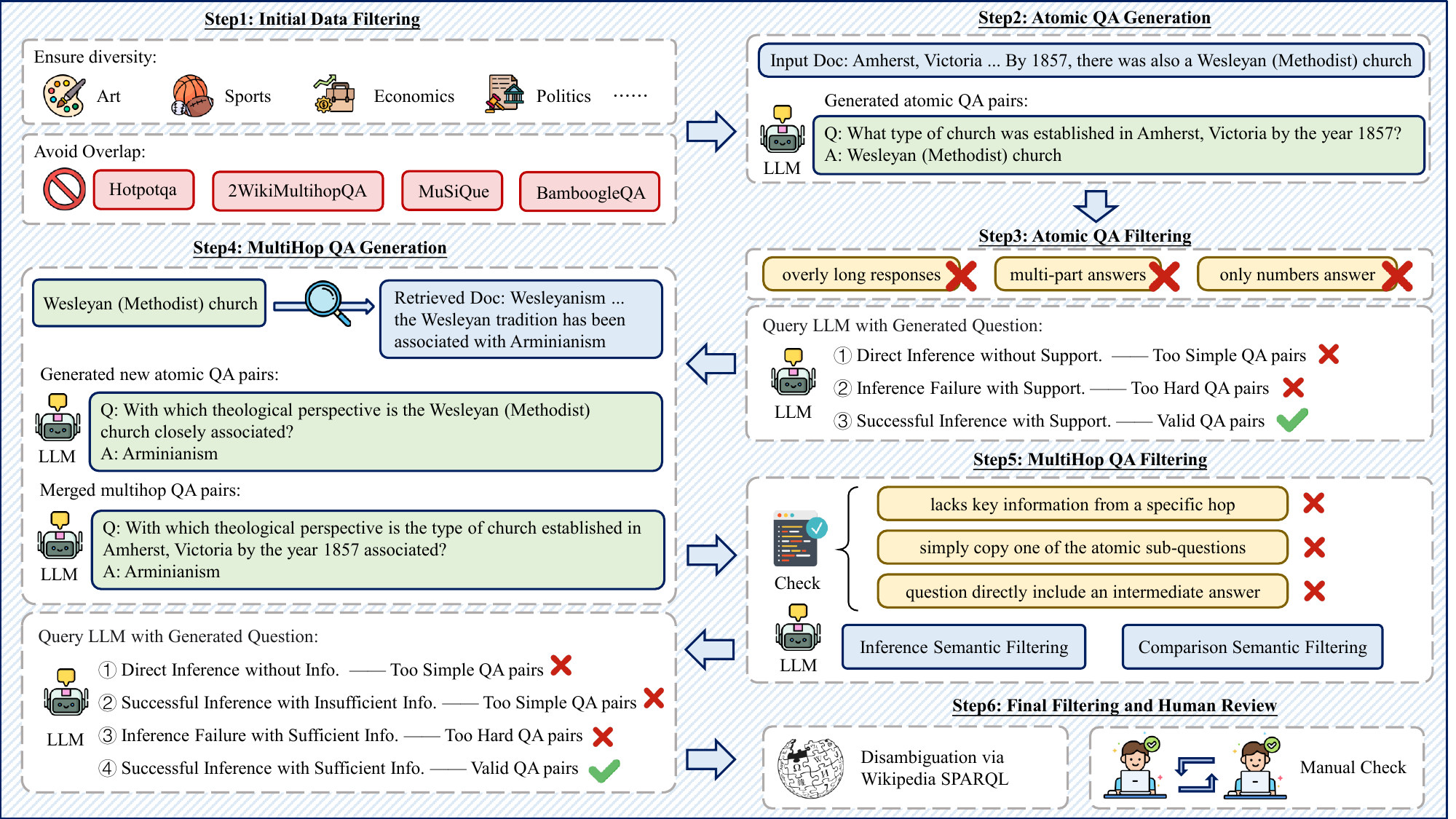}
    \caption{Data construction workflow.}
    \label{fig:data_construction}
    \vspace{-2mm}
\end{figure*}
Through a rigorous analysis of existing benchmarks (see Appendix~\ref{app:prior_work_case}), we identify four critical limitations that hinder the effective diagnosis of Agentic RAG systems: 
\textbf{Authenticity of Multi-hop Reasoning.} The complexity labels in existing datasets are often inflated. We find that questions labeled as multi-hop can frequently be answered directly by the model's internal knowledge or require fewer reasoning steps than claimed. This misalignment compromises the validity of reasoning evaluations. 
\textbf{Faithfulness to RAG Settings.} While the logical connections between hops may appear plausible to humans, the provided supporting documents often fail to supply explicit links. in such cases, the model is forced to implicitly rely on human prior knowledge rather than actual retrievable evidence, violating the fundamental premise of RAG systems. 
\textbf{Neglect of Intermediate Reasoning.} Most benchmarks provide only the final supporting documents (or sentences) without the complete reasoning trajectory. For modern agents equipped with retrieval tools, errors often occur at specific intermediate steps (e.g., query formulation or document selection). Current benchmarks fail to capture which step fails, treating the agent as a black box. 
\textbf{Lack of Knowledge Bases and Indices.} Existing benchmarks typically do not release the specific corpus and retrieval index used during construction. This omission not only makes evaluation cumbersome but also severely hinders reproducibility, as retrieval performance is highly sensitive to the underlying index construction.

To address these limitations, we propose \benchmarkname, a Hop-Aware benchmark specifically designed for diagnosing multi-step retrieval reasoning in Agentic RAG. We design an automated pipeline to generate high-quality questions with clear topological structures: Inference (Sequential) and Comparison (Parallel). By simulating the interactive workflow of agents, we iteratively combine atomic questions and apply rigorous filtering with final human verification. Evaluation results on mainstream LLMs reveal significant challenges in multi-hop reasoning, highlighting the need for this diagnostic tool.

Our main contributions are summarized as follows:
\begin{itemize}
    \item We propose \benchmarkname, a benchmark consisting of 1,305 examples, all of which are rigorously human-verified to ensure logical integrity. To the best of our knowledge, it is the first benchmark to enable hop-aware analysis, providing the granular evidence necessary to diagnose model performance at each specific reasoning and retrieval step.
    \item We design an automatic pipeline for constructing high-quality multi-hop Agentic RAG data, featuring a multi-stage verification protocol that blends automated logic filtering with human-in-the-loop adjudication. This framework enables the scalable and convenient generation of diverse, high-fidelity datasets spanning multiple domains.
    \item Through a comprehensive evaluation of 13 mainstream models, we demonstrate that existing LLMs—even state-of-the-art systems—exhibit severe degradation on complex multi-hop agentic reasoning. Leveraging hop-aware analysis, our benchmark uncovers how errors accumulate across reasoning steps, specifically how the mismatch between reasoning steps and task structure leads to collapsed or over-extended trajectories that diverge from the required logical progression, highlighting limitations that are not captured by conventional end-task evaluations.
\end{itemize}

\begin{figure*}[h]
    \centering
    \includegraphics[width=1\linewidth]{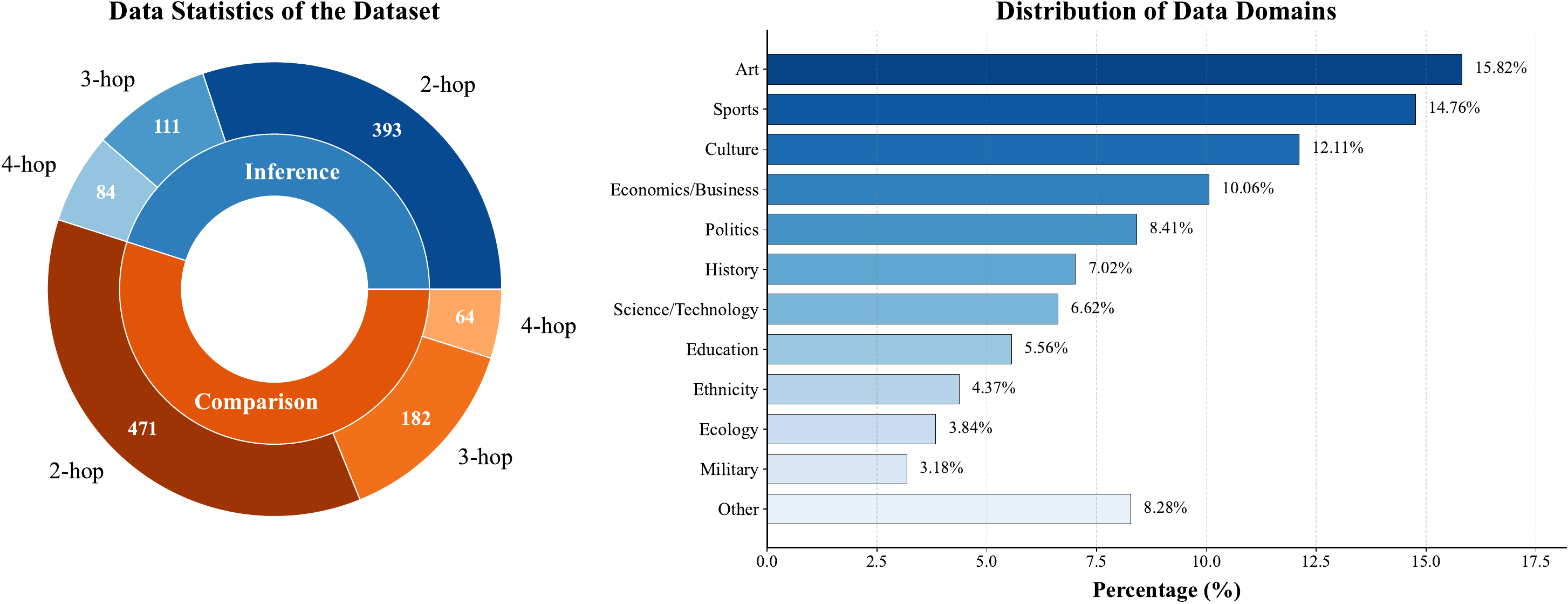}
    \caption{Data Statistics}
    \label{fig:data_statistics}
\end{figure*}

\section{Related Work}

\subsection{Agentic RAG}
Retrieval-Augmented Generation (RAG) was originally introduced to combine sequence generation with retrieval, allowing models to leverage external knowledge~\citep{rag}. Subsequent research developed Advanced~\citep{advanced}, Modular~\citep{modular}, and Graph RAG~\citep{graph} systems, which improved performance but largely operated as static pipelines.
Recently, the field has shifted towards \textbf{Agentic RAG}, which integrates autonomous agent capabilities—such as planning, tool use, and reflection—into the retrieval framework. Recent studies formalize these systems, highlighting how agents can dynamically decide what to retrieve and orchestrate multi-step interactions~\citep{agentic, liang2025reasoningrag12}. Parallel efforts explore reinforcement learning to optimize these reasoning-search trajectories~\citep{search-r1, research}, transforming RAG from a static process into an adaptive, goal-oriented workflow. For more advanced data-centric agenticrag approaches, please refer to ~\citet{deng2025data}.

\subsection{Benchmarks for Multi-hop Retrieval}
With the rapid development of Agentic RAG in recent years, there is a growing need for benchmarks that can effectively evaluate its performance. Multi-hop questions are particularly suitable for this purpose, as they require logical decomposition and the use of multiple documents, making them a natural testbed for assessing agent capabilities. Currently, most evaluations still rely on earlier benchmarks created through manual synthesis, such as HotpotQA, 2WikiMultihopQA, MuSiQue, and BamboogleQA. However, the construction of these datasets is labor-intensive, as it requires manually consulting large numbers of documents, and they also suffer from several inherent limitations.  

More recently, benchmarks supported by LLM-assisted question generation have been proposed, including MoreHopQA \citep{morehopqa}, MultiHop-RAG \citep{multihop-rag}, and MINTQA \citep{mintqa}. Yet, these approaches also face shortcomings: MoreHopQA simply extends the hop length of existing benchmarks, while MultiHop-RAG and MINTQA rely on directly prompting LLMs to generate questions from raw Wikipedia documents, without ensuring fine-grained quality control. 

\subsection{Data Synthesis}
Recently, data synthesis has emerged as an important technique for improving the performance of large language models (LLMs)~\cite{JCST-2509-15948, bai2024survey, an2025unictokens, guo2025video, lin2025perceiveanythingrecognizeexplain, an2024mc, luo2024llm}. Prior work has extensively explored data synthesis for both textual and multimodal domains. In the text domain, LLM-driven data synthesis pipelines are typically constructed using complex, workflow-based systems such as DataFlow~\cite{liang2025dataflow, cai2025text2sql, shen2025let, zheng2024pas, liang2024synth}, enabling high-quality synthetic data generation and achieving strong performance across a wide range of downstream tasks. In the multimodal domain, data synthesis has also proven effective. For example, prior studies synthesize large-scale image caption datasets~\cite{liu2024synthvlm} or multimodal verification trajectories~\cite{sun2025mm} to enhance the training and reasoning capabilities of vision-language models.

\section{Data Construction}

\subsection{Initial Data Filtering}
As shown in Figure~\ref{fig:data_construction}, we first randomly sampled a collection of documents from the Wikipedia dump\footnote{Given that numerous recent Agentic RAG training efforts have adopted the index and corpus released by FlashRAG, we follow the same collection to construct our benchmark, \url{https://www.modelscope.cn/datasets/hhjinjiajie/FlashRAG_Dataset} thereby ensuring a fair and comparable evaluation of model performance.}. To avoid overlap with existing benchmarks (including HotpotQA, 2Wiki, and MuSiQue), we compared their titles against those provided in the existing datasets and removed duplicates. We then employed an LLM \footnote{by default GPT-4o-mini, unless otherwise specified in the paper} to annotate their types. This step prevents an overrepresentation of any single question type and ensures diversity within our benchmark.

\subsection{Atomic Question Generation}
We synthesized atomic QA pairs from the filtered Wikipedia documents using an LLM, followed by a rigorous quality assurance pipeline. The prompts used for data construction are provided in Appendix ~\ref{app:prompt used}. First, we applied heuristic filtering to remove malformed responses; crucially, we excluded answers consisting solely of numbers to preserve semantic context, thereby preventing relevance drift in subsequent retrieval hops. Next, to ensure the necessity of retrieval, we discarded questions that the LLM could correctly answer using only its parametric knowledge. Finally, we verified grounding solvability by re-prompting the LLM with the source document, retaining only those questions where the provided context was sufficient to derive the correct answer.

\subsection{Design of Multi-hop Questions}
To comprehensively evaluate agentic reasoning, we design two distinct question topologies: \textbf{Sequential (Inference)}, which requires a stepwise logical chain where intermediate answers bridge to the next step; and \textbf{Parallel (Comparison)}, which necessitates gathering independent information about multiple entities before synthesizing a final answer. 

\noindent\textbf{Construction of 2-hop Questions.} We synthesize 2-hop queries by retrieving additional documents relevant to our atomic QA pairs, explicitly excluding documents present in existing benchmarks to prevent leakage. An LLM is then prompted to combine original and retrieved atomic pairs into complex queries, dynamically selecting between inference or comparison templates. Each candidate question undergoes a rigorous three-stage verification protocol:

\noindent\textbf{1) Structural Integrity Filtering.} We discard candidates with obvious syntactic or logical flaws. Crucially, we filter out questions exhibiting \textit{information leakage} (where intermediate answers are explicitly revealed in the question text) and those resulting from \textit{trivial concatenation} (where sub-questions are simply joined without semantic integration).

\noindent\textbf{2) Semantic Logic Verification.} An LLM auditor is employed to reject incoherent reasoning chains. For \textit{inference questions}, we eliminate spurious links, such as forcefully connecting unrelated entities or conflating distinct entities with similar names. For \textit{comparison questions}, we discard invalid comparisons between dissimilar concepts or those lacking a clear comparative basis.

\noindent\textbf{3) Multi-hop Necessity \& Solvability Check.} To guarantee the validity of the reasoning chain, we enforce three criteria: (i) \textit{Retrieval Necessity}: The question must not be answerable by the model's parametric memory alone; (ii) \textit{Dependency Irreducibility}: Given the full document set, removing \textit{any single} supporting document must render the question unanswerable; (iii) \textit{Grounded Solvability}: The model must successfully derive the correct answer when provided with the complete evidence set.

\noindent\textbf{Extension to 3--4 Hop Questions.} To challenge the limits of sequential and parallel planning, we iteratively extend verified 2-hop questions by appending additional inference steps. These higher-order questions undergo the same stringent filtering pipeline described above to ensure they remain natural and logically solvable, yielding the final benchmark.

\begin{table*}[!htbp]
\centering
\renewcommand{\arraystretch}{1.1}
\caption{Results of Closed-Sourced and Open-Sourced LLMs on our AgenticRAGTracer.}
\label{tab:multihop}
\adjustbox{max width=\textwidth}{%
\begin{tabular}{c|ccc|ccc|ccc|ccc|ccc|ccc}
\toprule
\multirow{4}{*}{\textbf{Models}}
& \multicolumn{9}{c|}{\textbf{Comparison}} 
& \multicolumn{9}{c}{\textbf{Inference}} \\
\cmidrule(lr){2-10}\cmidrule(lr){11-19}
& \multicolumn{3}{c|}{2hop} & \multicolumn{3}{c|}{3hop} & \multicolumn{3}{c|}{4hop}
& \multicolumn{3}{c|}{2hop} & \multicolumn{3}{c|}{3hop} & \multicolumn{3}{c}{4hop} \\
\cmidrule(lr){2-4}\cmidrule(lr){5-7}\cmidrule(lr){8-10}
\cmidrule(lr){11-13}\cmidrule(lr){14-16}\cmidrule(lr){17-19}
& EM & F1 & LLM & EM & F1 & LLM & EM & F1 & LLM
& EM & F1 & LLM & EM & F1 & LLM & EM & F1 & LLM \\
\midrule

\rowcolor{blue!15}
\multicolumn{19}{c}{\textbf{Closed-Sourced LLMs}} \\
\midrule
\textbf{o4-mini} &77.7 &85.1 &86.0 &67.6 &74.0 &79.1 &\cellcolor{gray!20}\textbf{35.9} &\cellcolor{gray!20}\textbf{47.7} &\cellcolor{gray!20}\textbf{54.7}
&47.8 &57.2 &66.2 &\cellcolor{gray!20}\textbf{41.4} &\cellcolor{gray!20}\textbf{45.4} &\cellcolor{gray!20}\textbf{48.7} &21.4 &31.8 &34.5 \\

\textbf{GPT-5} &76.2 &85.4 &88.1 &\cellcolor{gray!20}\textbf{69.8} &\cellcolor{gray!20}\textbf{76.0} &81.3 &26.6 &39.3 &45.3
&\cellcolor{gray!20}\textbf{48.4} &\cellcolor{gray!20}\textbf{59.0} &\cellcolor{gray!20}\textbf{70.0} &33.3 &40.3 &42.3 &\cellcolor{gray!20}\textbf{22.6} &\cellcolor{gray!20}\textbf{34.0} &\cellcolor{gray!20}\textbf{39.3} \\

\textbf{GPT-4o} &69.2 &76.3 &75.2 &51.7 &58.5 &59.9 &15.6 &23.8 &20.3 &24.9 &33.0 &35.1 &13.5 &18.4 &17.1 &4.8 &10.0 &7.1 \\

\textbf{Grok-4} &\cellcolor{gray!20}\textbf{80.7} &\cellcolor{gray!20}\textbf{87.0} &\cellcolor{gray!20}\textbf{89.0} &66.5 &73.8 &\cellcolor{gray!20}\textbf{81.9} &32.8 &46.9 &53.1
&41.7 &51.9 &61.6 &34.2 &40.4 &40.5 &20.2 &28.9 &31.0 \\

\textbf{Grok-3} &75.6 &84.3 &83.4 &63.7 &70.8 &75.8 &23.4 &37.7 &39.1
&34.6 &43.4 &50.4 &19.8 &24.4 &24.3 &4.8 &10.8 &10.7 \\
\midrule

\rowcolor{yellow!15}
\multicolumn{19}{c}{\textbf{Open-Sourced LLMs}} \\
\midrule
\textbf{DeepSeek-R1} &74.1 &82.0 &81.7 &43.4 &52.7 &59.9 &15.6 &24.3 &32.8 &27.7 &37.7 &43.3 &16.2 &21.8 &20.7 &4.8 &10.7 &7.1 \\

\textbf{DeepSeek-V3} &66.0 &72.1 &72.2 &37.4 &41.7 &44.0 &15.6 &23.7 &23.4 &22.9 &30.1 &33.3 &10.8 &15.3 &15.3 &2.4 &7.5 &6.0 \\

\textbf{Qwen2.5-72B} &78.8 &85.0 &84.3 &62.1 &73.2 &\cellcolor{gray!20}\textbf{81.9} &21.9 &34.1 &35.9 &37.7 &46.8 &53.4 &18.0 &25.1 &26.1 &8.3 &14.3 &11.9 \\

\textbf{Qwen2.5-32B} &70.7 &77.5 &76.7 &50.6 &60.3 &64.8 &18.8 &26.8 &26.6 &34.6 &42.4 &48.4 &11.7 &14.5 &16.2 &6.0 &9.1 &9.5 \\

\textbf{Qwen2.5-7B} &52.4 &65.1 &70.5 &36.8 &44.0 &47.3 &7.8 &16.1 &9.4 &24.7 &32.8 &35.1 &7.2 &13.1 &13.5 &2.4 &8.2 &4.8 \\

\textbf{R-Search} &72.2 &80.9 &78.3 &39.0 &47.7 &51.7 &14.1 &22.3 &15.6 &26.0 &36.0 &41.5 &13.5 &23.8 &20.7 &7.1 &13.0 &9.5 \\

\textbf{Search-R1} &64.5 &74.5 &65.4 &44.5 &53.9 &48.9 &15.6 &25.7 &12.5 &25.5 &34.8 &37.4 &11.7 &16.4 &14.4 &6.0 &9.5 &7.1 \\

\textbf{ReSearch} &60.3 &69.9 &67.7 &47.3 &56.6 &55.0 &12.5 &25.7 &21.9 &28.2 &37.5 &41.2 &17.1 &20.3 &20.7 &4.8 &9.2 &7.1 \\
\bottomrule
\end{tabular}
}
\end{table*}
\subsection{Human Review and Quality Control}  
In the final stage, we apply Wikipedia SPARQL queries to remove potentially ambiguous cases where multiple entities share the same name in Wikipedia. Following this automated disambiguation, to strictly guarantee data quality, we conducted a full-scale manual verification of the entire dataset. 

Specifically, each generated data were independently evaluated by three trained human annotators. The evaluation was formulated as a binary decision (Retain or Discard) driven by two primary dimensions: (1) general quality checks to ensure strict factuality, faithfulness to the source text, and grammatical fluency; and (2) type-specific checks to validate the validity of intermediate reasoning steps and the dimensional consistency for comparison questions. 

To assess the reliability of the human evaluation, we calculated Fleiss' Kappa ($\kappa$), achieving a score of 0.65, which indicates substantial inter-annotator agreement. For non-unanimous cases, the final labels were rigorously determined through a consensus discussion involving both the annotators and the authors. Further details are provided in Appendix \ref{app:human_eval}.

\begin{table*}[t]
\centering
\caption{Reasoning depth and average steps on AgenticRAGTracer.
MaxD denotes the maximum reasoning depth reached by each model.
C-Steps and I-Steps denote the average number of steps on correctly and incorrectly answered samples, respectively.}
\label{tab:xxx}
\adjustbox{max width=\textwidth}{%
\begin{tabular}{c|ccc|ccc|ccc|ccc}
\toprule
\multirow{3}{*}{\textbf{Models}} &
\multicolumn{6}{c|}{\textbf{Comparison}} &
\multicolumn{6}{c}{\textbf{Inference}} \\
\cmidrule(lr){2-7}\cmidrule(lr){8-13}
& \multicolumn{3}{c|}{3hop} & \multicolumn{3}{c|}{4hop}
& \multicolumn{3}{c|}{3hop} & \multicolumn{3}{c}{4hop} \\
\cmidrule(lr){2-4}\cmidrule(lr){5-7}
\cmidrule(lr){8-10}\cmidrule(lr){11-13}
& MaxD & Steps-C & Steps-I
& MaxD & Steps-C & Steps-I
& MaxD & Steps-C & Steps-I
& MaxD & Steps-C & Steps-I \\
\midrule

\rowcolor{blue!15}
\multicolumn{13}{c}{\textbf{Closed-Sourced LLMs}} \\
\midrule
\textbf{o4-mini}      & 2.62 & 3.02 & 5.87 & 3.23 & 4.34 & 6.03 & 1.77 & 3.02 & 5.87 & 2.11 & 4.72 & 7.78 \\
\textbf{GPT-5}        & 2.70 & 3.13 & 5.85 & 3.20 & 4.10 & 7.03 & 1.79 & 3.14 & 5.85 & 2.40 & 4.48 & 8.25 \\
\textbf{GPT-4o}       & 2.43 & 3.15 & 3.38 & 2.81 & 4.08 & 4.18 & 0.96 & 3.16 & 3.38 & 1.02 & 3.83 & 4.92 \\
\textbf{Grok-4}       & 2.70 & 3.19 & 3.76 & 3.16 & 4.65 & 5.57 & 1.65 & 3.19 & 3.76 & 2.00 & 4.69 & 6.95 \\
\textbf{Grok-3}       & 2.53 & 3.01 & 3.00 & 3.00 & 4.44 & 4.62 & 1.09 & 3.01 & 3.00 & 0.94 & 4.44 & 5.47 \\

\midrule
\rowcolor{yellow!15}
\multicolumn{13}{c}{\textbf{Open-Sourced LLMs}} \\
\midrule
\textbf{DeepSeek-R1}  & 2.38 & 3.04 & 3.26 & 2.94 & 4.10 & 3.28 & 1.04 & 3.05 & 3.26 & 0.93 & 4.17 & 2.76 \\
\textbf{DeepSeek-V3}  & 2.07 & 3.08 & 2.81 & 2.53 & 4.07 & 3.49 & 0.86 & 3.09 & 2.81 & 0.83 & 4.20 & 3.61 \\
\textbf{Qwen2.5-72B}  & 2.76 & 3.28 & 3.54 & 3.03 & 4.30 & 4.41 & 1.16 & 3.28 & 3.55 & 0.94 & 4.50 & 7.03 \\
\textbf{Qwen2.5-32B}  & 2.49 & 3.08 & 2.83 & 2.46 & 4.18 & 3.51 & 1.03 & 3.08 & 2.83 & 0.83 & 4.13 & 2.96 \\
\textbf{Qwen2.5-7B}   & 2.20 & 2.95 & 3.04 & 2.52 & 4.50 & 3.41 & 0.93 & 3.01 & 3.04 & 0.89 & 5.00 & 3.66 \\
\textbf{R-Search}    & 2.32 & 3.08 & 2.97 & 2.78 & 4.20 & 3.65 & 0.96 & 3.09 & 2.97 & 0.93 & 4.13 & 3.57 \\
\textbf{Search-R1}     & 2.28 & 3.03 & 2.83 & 2.27 & 4.38 & 3.88 & 1.03 & 3.03 & 2.83 & 0.90 & 4.17 & 4.50 \\
\textbf{ReSearch}     & 2.26 & 3.26 & 2.99 & 2.58 & 4.21 & 3.88 & 1.09 & 3.26 & 2.99 & 0.85 & 4.00 & 4.06 \\

\bottomrule
\end{tabular}
}
\end{table*}
\subsection{Data statistics}
As shown in Figure~\ref{fig:data_statistics}, our benchmark exhibits both semantic diversity and structural complexity. The domain distribution (right) spans over 11 categories ranging from Art to Military, with no single domain dominating more than 16\% of the dataset, ensuring a balanced evaluation. The question topology (left) features a balanced composition of Inference and Comparison types, facilitating a comprehensive assessment of both sequential and parallel reasoning capabilities. Moreover, the dataset spans a difficulty spectrum from fundamental 2-hop queries to complex 3- and 4-hop challenges. This progressive design allows for a fine-grained analysis, stepwise unveiling the capability boundaries of agentic models. 
Due to space constraints, representative examples of our benchmark are provided in Appendix~\ref{app:classification data statistics} to illustrate its structure and complexity.

\section{Experiments}
\subsection{Setup}
We evaluate a diverse set of LLMs on our benchmark, including closed-source models (GPT and Grok series), open-source models (DeepSeek \citep{deepseek-v3, deepseek-r1} and Qwen \citep{qwen} series), as well as checkpoints of recent Agentic RAG models trained from Qwen2.5-7B-Instruct (Search-R1 \citep{search-r1}, ReSearch \citep{research}, and R-Search \citep{r-search}).
To ensure a fair comparison, all models are evaluated under a unified experimental setting using the Qwen-Agent framework, which follows the ReAct \citep{react} paradigm to interleave reasoning and actions.
We report Exact Match (EM), F1 score, and LLM-as-a-Judge scores, combining objective metrics with model-based evaluation.
Regarding the LLM-as-a-Judge scores, we employ GPT-4o-mini with the temperature set to 0 to ensure deterministic and reproducible results. To verify the rationality and stability of this protocol, we cross-checked the judge's verdicts on a random sample (20 instances per subset) against stronger models—specifically GPT-5 and Grok-4, as well as human annotations. The results demonstrate near-perfect agreement across these evaluators, confirming that the LLM judge provides a reliable and unbiased measure of model performance. The detailed prompts used for the evaluation are provided in the Appendix ~\ref{app:prompt used}.

\subsection{Main Results}
The primary results of our evaluation on the proposed benchmark are detailed in Table~\ref{tab:multihop}. We summarize the main findings below:

Overall, the benchmark poses a challenge to LLM's multi-hop reasoning capability. Even GPT-5 only has a 22.6\% EM score on the 4-hop inference benchmark.

Across all models, performance declines as the number of reasoning hops increases from two to four across both Comparison and Inference subsets; however, the magnitude of this degradation varies substantially across models, revealing different levels of robustness to long-chain reasoning.
Similarly, while all models perform worse on the Inference subset, the resulting performance gap differs markedly, indicating that the benchmark effectively differentiates models in their ability to handle serial reasoning and dependency tracking.

Closed-source models such as GPT-5, o4-mini, and Grok-4 generally outperform most open-source counterparts, demonstrating their stronger integrated reasoning and retrieval coordination abilities. However, an exception is GPT-4o, which performs unexpectedly poorly on our benchmark. To investigate the underlying causes of this performance discrepancy, we conducted a quantitative analysis of the retrieval behavior across different models, specifically examining the retrieval volume per step.

\begin{figure}[t]
\centering
\includegraphics[width=1\linewidth]{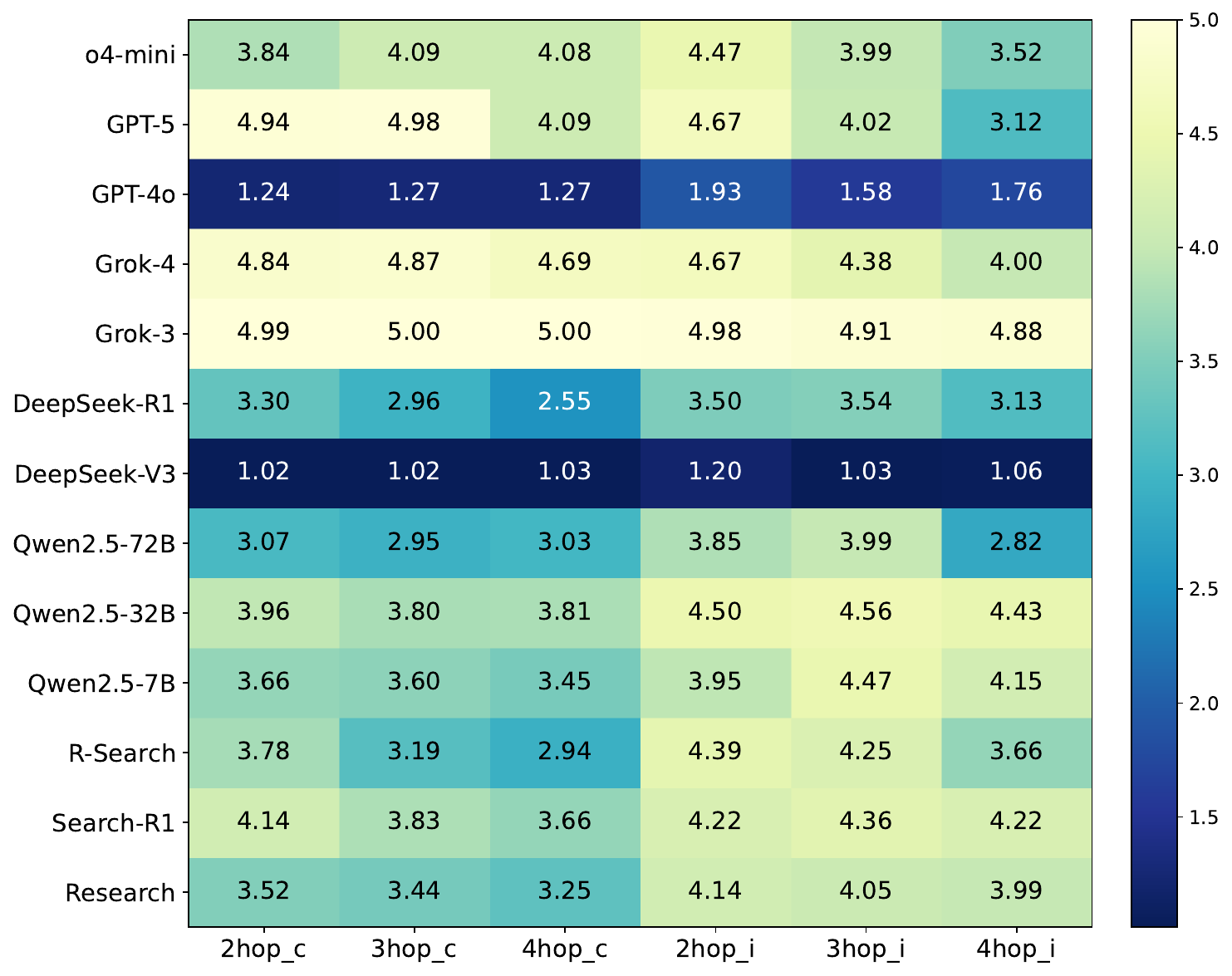}
\caption{Average top-k values.}
\label{fig:topk_stats}
\end{figure}

As illustrated in Figure~\ref{fig:topk_stats}, there is a distinct correlation between the retrieval strategy and overall performance. Top-tier models like Grok-4 and GPT-5 consistently employ larger top-k values, reflecting a strategy of broad evidence integration. In the context of agentic RAG, this "information-seeking" behavior is crucial, as missing a single piece of evidence in the early hops of a multi-step chain can irreversibly derail the entire reasoning process.

In stark contrast, GPT-4o adopts a highly conservative retrieval strategy, consistently retrieving far fewer documents per step—a behavior pattern notably similar to DeepSeek-V3. This restrictive approach creates an information bottleneck, resulting in limited information accumulation that fails to satisfy the grounded solvability requirements of complex multi-hop queries. Consequently, even with strong internal reasoning capabilities, the model fails because it operates on an incomplete context. This finding not only explains GPT-4o's degradation but also highlights a key diagnostic capability of our benchmark: it effectively exposes suboptimal retrieval strategies (e.g., over-confidence or premature stopping) that limit multi-hop reasoning effectiveness, which traditional single-hop benchmarks may overlook.

\begin{figure*}[h]
    \centering
    \includegraphics[width=1\linewidth]{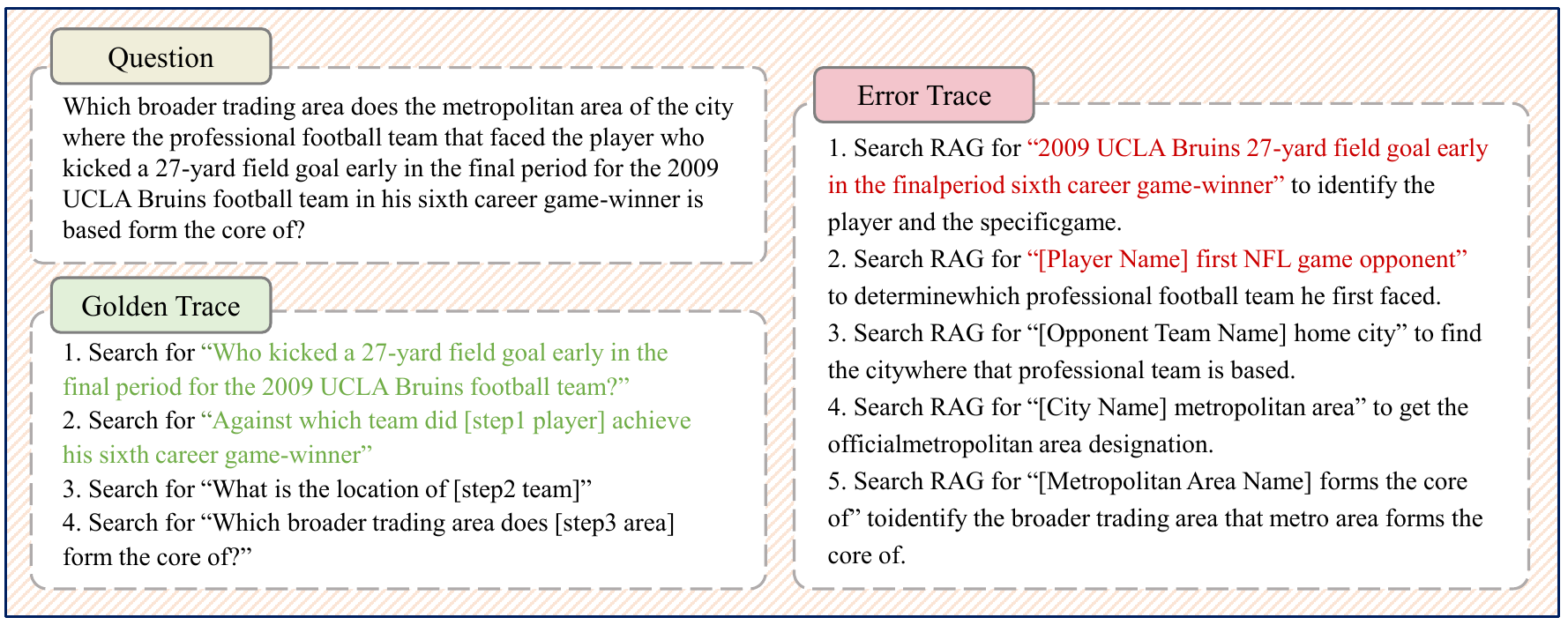}
    \caption{Error case study. We present the key information in the figure; additional details can be found in Appendix~\ref{app:error_case}.}
    \label{fig:error_case_study}
\end{figure*}

Within open-source model families, performance scales positively with model size: Qwen2.5-72B-Instruct consistently outperforms its 32B and 7B variants, suggesting that larger capacity improves multistep reasoning stability. And Qwen2.5-72B-Instruct outperforms the DeepSeek series, indicating a stronger tool-calling and information retrieval capability. Moreover, checkpoints derived from Qwen2.5-7B-Instruct substantially outperform the base model; in some cases, they approach larger open-source models, though a gap to closed-source models remains.

\subsection{Multi-Hop Analysis}
To better understand model performance on multi-step reasoning, we examine the maximal reasoning depth that models can handle when failing on higher-hop questions. For each 3- and 4-hop question, we record the average of the highest corresponding lower-hop questions that the model answers correctly (Table \ref{tab:xxx}). This metric captures the depth of reasoning that a model can achieve, even when it fails on longer chains.
This analysis serves two purposes. First, it highlights the limitations of the current models in maintaining coherent, iterative reasoning: frequent failures on high-hop questions, despite solving the initial hops, indicate weaknesses in multi-step integration. Second, it validates our benchmark design: by quantifying performance degradation across hop levels, \benchmarkname demonstrates its ability to stress agentic reasoning beyond end-task accuracy. Instead of reporting only overall accuracy, this cross-hop evaluation provides a layered, interpretable view of reasoning behavior, revealing that a model may succeed in 2-hop reasoning but struggle with 3- or 4-hop chains, which is unattainable with traditional benchmarks.

We also analyze the relationship between the reasoning step count and answer correctness on the 3- and 4-hop subsets (Table \ref{tab:xxx}), which represents the challenging and representative setting of our benchmark. Across all models, correct responses exhibit step counts closely aligned with the target hop number, indicating structurally consistent reasoning trajectories. In contrast, incorrect responses show significantly more diverse behavior, with both over-extended and prematurely terminated reasoning chains. This gap reveals that success in multihop RAG depends not on longer reasoning but on allocating an appropriate number of steps consistent with task structure, which is a distinction enabled by our benchmark but obscured in prior evaluations.

\subsection{Error Case Study}
To gain a deeper understanding of model failures, we conducted a systematic analysis of the agentic reasoning trajectories. Our findings reveal that the vast majority of failures are not isolated errors in tool usage but rather cascading failures fundamentally rooted in a breakdown of initial task decomposition (Figure~\ref{fig:error_case_study}). As illustrated in the figure, the LLM incorrectly placed the specific token \textit{'sixth career game-winner'} into the first reasoning step; subsequently, it introduced a spurious subgoal—\textit{'the professional football team he first faced'}—which lacked a logical basis in the original query. This initial analytical misstep propagated a chain of errors throughout the reasoning trajectory, ultimately rendering the final inference completely incorrect.

The core bottleneck lies in the model's inability to correctly partition a complex, multi-hop query into a valid logical sequence. A single analytical misstep at this stage immediately diverts the agent onto an erroneous reasoning trajectory. Once the agent enters such a misaligned path, its entire subsequent interactive process becomes fundamentally unproductive; the agent effectively "loses its way," pursuing a flawed objective where even successful retrieval steps fail to contribute to the correct answer. In this state of trajectory deviation, the model’s efforts—such as issuing redundant queries or misinterpreting retrieved context—are merely inevitable byproducts of its commitment to a faulty plan. Ultimately, these observations underscore that the reliability of Agentic RAG is primarily governed by the robustness of its high-level logical planning, where the initial hop determines the viability of the entire reasoning chain.

\subsection{Discussion}
The empirical findings from AgenticRAGTracer reveal a fundamental gap between "following a planning prompt" and "possessing true reasoning agency." Although our evaluation pipeline mandates an initial planning stage, the high failure rates on complex tasks—coupled with the cascading errors illustrated in Figure 3—highlight a pervasive rigidity in plan execution. Most models treat their initial decomposition not as a flexible working hypothesis, but as an immutable script. When a model encounters a dead end or retrieves information that contradicts its initial assumptions, it lacks the autonomous meta-cognition to pause and re-evaluate its trajectory. This reasoning inertia is quantitatively evidenced in Table 2, where the reasoning steps for failed tasks (Steps-I) significantly exceed those for successful ones (Steps-C). This "flailing" behavior—characterized by redundant queries and over-extension—suggests that current agents cannot autonomously distinguish between a path that requires more evidence and one that has entered a logical abyss. True autonomous agency in RAG requires more than just better planning; it demands a dynamic self-audit mechanism capable of proactive course correction and informed termination. Our benchmark serves as a crucial diagnostic tool in this regard, proving that the bottleneck in multi-hop reasoning is not merely the lack of information, but the inability to strategically and autonomously manage the reasoning process itself.

\section{Conclusion}
In this paper, we introduce AgenticRAGTracer, the first hop-aware benchmark for evaluating Agentic RAG systems, together with an automated pipeline for constructing high-quality multi-hop questions. By iteratively synthesizing atomic facts into complex Inference and Comparison topologies, our pipeline ensures that each question possesses a clear, traceable reasoning structure backed by rigorous logical verification. Extensive experiments reveal that model failures are predominantly characterized by distorted reasoning chains, either collapsing prematurely or drifting into over-extension, reflecting an inability to allocate reasoning steps in alignment with the logical structure of the task.
AgenticRAGTracer contributes a novel diagnostic perspective by shifting the focus toward step-level transparency. A core strength of our work lies in the high logical fidelity of our benchmark; by combining a controlled automated pipeline with rigorous human verification, we ensure a reliable and traceable dataset that goes beyond simple end-to-end metrics. We hope this resource provides a useful foundation for the community to identify subtle failure modes and encourages the development of more resilient, self-correcting Agentic RAG systems.

\bibliography{latex/custom}
\appendix
\clearpage
\begin{figure*}[!htbp]
\section{Data Example}
\label{app:classification data statistics}
\centering
\small
\captionof{table}{Data Example. To make it look clearer, We only present the key information from the supporting documents that is directly used in question generation.}
\label{data_example}
\begin{tabular}{l c p{12cm}}
\toprule
Inference   & 2-hop & 1-hop Question: \textcolor{blue}{What type of church was established in Amherst, Victoria by the year 1857?} \\
            & & 1-hop Answer: Wesleyan (Methodist) church \\
            & & 1-hop Doc: Amherst, Victoria ... By 1857, there was also a Wesleyan (Methodist) church ... \\
            & & New Atomic Question: \textcolor{Green}{With which theological perspective is the Wesleyan (Methodist) church closely associated?} \\
            & & New Atomic Answer: Arminianism \\
            & & New Doc: Wesleyanism ... the Wesleyan tradition has been associated with Arminianism ... \\
            & & 2-hop Question: \textcolor{Green}{With which theological perspective} is the type of \textcolor{blue}{church established in Amherst, Victoria by the year 1857} associated? \\
            & & 2-hop Answer: Arminianism \\
            & 3-hop & New Atomic Question: \textcolor{orange}{Who is the Dutch Reformed theologian associated with the concept of Arminianism?} \\
            & & New Atomic Answer: Jacobus Arminius \\
            & & New Doc: Arminianism ... is based on theological ideas of the Dutch Reformed theologian Jacobus Arminius ... \\
            & & 3-hop Question: \textcolor{orange}{Who is the Dutch Reformed theologian} \textcolor{Green}{associated with the theological perspective of} \textcolor{blue}{the church established in Amherst, Victoria by the year 1857?} \\
            & & 3-hop Answer: Jacobus Arminius \\
            & 4-hop & New Atomic Question: \textcolor{Goldenrod}{Where did Jacobus Arminius teach his sermons on the Epistle of the Romans?} \\
            & & New Atomic Answer: Amsterdam \\
            & & New Doc: Jacobus Arminius ... At Amsterdam, Arminius taught through "a number of sermons on the Epistle of the Romans." ... \\
            & & 4-hop Question: \textcolor{Goldenrod}{Where did} \textcolor{orange}{the Dutch Reformed theologian associated with}  \textcolor{Green}{the theological perspective of} \textcolor{blue}{the church established in Amherst, Victoria by the year 1857} \textcolor{Goldenrod}{teach his sermons on the Epistle of the Romans?} \\
            & & 4-hop Answer: Amsterdam \\
\midrule
Comparison  & 2-hop & 1-hop Question: \textcolor{blue}{What is the effective temperature of Omega Persei?} \\
            & & 1-hop Answer: 4,586 K \\
            & & 1-hop Doc: Omega Persei ... at an effective temperature of 4,586 K ... \\
            & & New Atomic Question: \textcolor{Green}{What is the effective temperature of the stellar atmosphere of HD 195564?} \\
            & & New Atomic Answer: 5,421 K \\
            & & New Doc: HD 195564 ...  The effective temperature of the stellar atmosphere is 5,421 K ... \\
            & & 2-hop Question: Which star has a higher effective temperature, \textcolor{blue}{Omega Persei} or \textcolor{Green}{HD 195564}? \\
            & & 2-hop Answer: HD 195564 \\
            & 3-hop & New Atomic Question: \textcolor{orange}{What spacecraft provided parallax measurements for HD 195564?} \\
            & & New Atomic Answer: Hipparcos \\
            & & New Doc: HD 195564 ... Parallax measurements from the Hipparcos spacecraft give us an estimate of its distance ... \\
            & & 3-hop Question: \textcolor{orange}{What spacecraft provided parallax measurements for the star} that has a higher effective temperature, \textcolor{blue}{Omega Persei} or \textcolor{Green}{HD 195564}? \\
            & & 3-hop Answer: Hipparcos \\
            & 4-hop & New Atomic Question: \textcolor{Goldenrod}{How many entries does the Hipparcos Catalogue contain?} \\
            & & New Atomic Answer: 118,218 entries \\
            & & New Doc: Hipparcos ... contains 118,218 entries (stars or multiple stars) ... \\
            & & 4-hop Question: \textcolor{Goldenrod}{How many entries does the catalogue produced by} \textcolor{orange}{the spacecraft that provided parallax measurements for the star} with a higher effective temperature between \textcolor{blue}{Omega Persei} and \textcolor{Green}{HD 195564} \textcolor{Goldenrod}{contain}? \\
            & & 4-hop Answer: 118,218 entries \\
            
\bottomrule
\end{tabular}
\end{figure*}

\FloatBarrier
\section{Case Studies of Limitations in Prior Work}
\label{app:prior_work_case}
The following examples illustrate the issues discussed above regarding prior work. \\
\textbf{Multi-hop Questions with Incorrect Step Counts:\\} 
\begin{table}[h] 
\centering
\renewcommand{\arraystretch}{1.1} 
\small
\begin{tabularx}{\linewidth}{>{\raggedright\arraybackslash}p{\linewidth}}
\toprule
\textbf{User:} \\
You are a world expert at making efficient plans to solve any task using an RAG Search tool. ...\\
Here is your task: What is the name of this White House Cheif of Staff, born in 1950, who assumed the role at the same time as Elaine Duke became Secretary of Homeland Security? \\
\midrule

\textbf{Assistant:} \\
1. Search “Elaine Duke Secretary of Homeland Security start date \\
2. Search “White House Chief of Staff assumed office on [date from step 1] \\
3. Search “[Name of Chief of Staff from step 2] born 1950 \\
\midrule

\textbf{User:} \\
Based on the plan and the search results before (if there is), first analyse what information you have gained and what other information you still need, then EXECUTE ONLY ONE MORE step using the RAG search tool. ... \\
\midrule

\textbf{Assistant:} \\
\texttt{"function\_call": \{"name": "RAG\_search", "arguments": "\{query: Elaine Duke Secretary of Homeland Security start date, topk: 3\}"\}} \\
\midrule

\textbf{[RAGTool Results]} \\
Query: "Elaine Duke Secretary of Homeland Security start date" \\
Topk = 3 \\
Document: \\
"Elaine Duke"
Elaine Duke Elaine Costanzo Duke (born June 26, 1958) is an American civil servant and former United States Deputy Secretary of Homeland Security, serving from April 10, 2017 until April 15, 2018. She became acting Secretary of Homeland Security on July 31, 2017, when John F. Kelly assumed the office of White House Chief of Staff. She left the acting position on December 6, 2017, upon the confirmation of Kirstjen Nielsen. The daughter of Francesco (Frank) Costanzo Duke and Concetta Scherma, Duke is a native of Ohio. Her maternal grandparents (Scherma) emigrated from Aidone, Sicily and her Costanzo family roots \\
\midrule

\textbf{User:} \\
Based on the plan and the search results before (if there is), first analyse what information you have gained and what other information you still need, then EXECUTE ONLY ONE MORE step using the RAG search tool. ... \\
\midrule

\textbf{Assistant: }\\
Final\_Answer: John F. Kelly \\

\bottomrule
\end{tabularx}
\caption{Authenticity Of multi-hop reasoning case}
\label{tab:authenticity_case}
\end{table}

\noindent The following presents the QA content from prior work (Hotpotqa): \\
\textbf{Question:} What is the name of this White House Cheif of Staff, born in 1950, who assumed the role at the same time as Elaine Duke became Secretary of Homeland Security? \\
\textbf{Answer: } John Francis Kelly \\
\textbf{Supporting\_Facts: } \\
title: Elaine Duke \\
Elaine Costanzo Duke (born 1958) is an American civil servant who is the current Acting United States Secretary of Homeland Security. \\
On January 30, 2017, she was nominated by President Donald Trump to become the United States Deputy Secretary of Homeland Security under John F. Kelly. \\
She became acting Secretary of Homeland Security on July 31, 2017, when John F. Kelly assumed the office of White House Chief of Staff. \\

\noindent \textbf{Unretrievable Questions for RAG: \\}
\textbf{Question:} Who is the spouse of the Green performer? \\
\textbf{Answer: } Miquette Giraudy \\
\textbf{supporting\_facts: } \\
Question: Green >> performer \\
Answer: Steve Hillage \\
Doc: Green is the fourth studio album by British progressive rock musician Steve Hillage. Written in spring 1977 at the same time as his previous album, the funk-inflected "Motivation Radio" (1977), "Green" was originally going to be released as "The Green Album" as a companion to "The Red Album" (the originally intended name for "Motivation Radio"). However, this plan was dropped and after a US tour in late 1977, "Green" was recorded alone, primarily in Dorking, Surrey, and in London. \\
Question: \#1 >> spouse \\
Answer: Miquette Giraudy \\
Doc: Miquette Giraudy (born 9 February 1953, Nice, France) is a keyboard player and vocalist, best known for her work in Gong and with her partner Steve Hillage. She and Hillage currently form the core of the ambient band System 7. In addition to her performances in music, she has also worked as an actress, film editor and writer. In each role, she has used different stage names. \\

\noindent In the above (from prior work Musique), it can be observed that the document does not contain any information about \emph{Miquette Giraudy spouse Green (Steve Hillage album)}, but only provides a brief description of Miquette Giraudy’s biography. As a result, the RAG tool fails to retrieve this information and consequently cannot obtain the correct answer.

\FloatBarrier
\section{Prompt Used}
\label{app:prompt used}
\begin{figure*}[!h]
    \centering
    \includegraphics[width=0.9\linewidth]{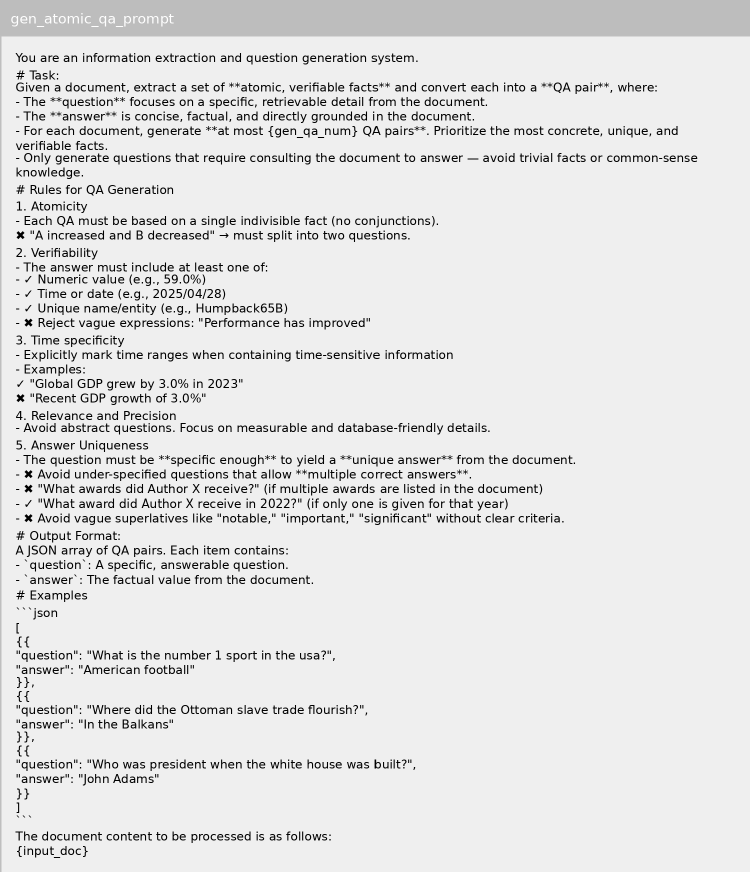}
    \caption{Gen atomic qa prompt}
    \label{fig:gen_atomic_qa_prompt}
\end{figure*}

\begin{figure*}[!h]
    \centering
    \includegraphics[width=0.9\linewidth]{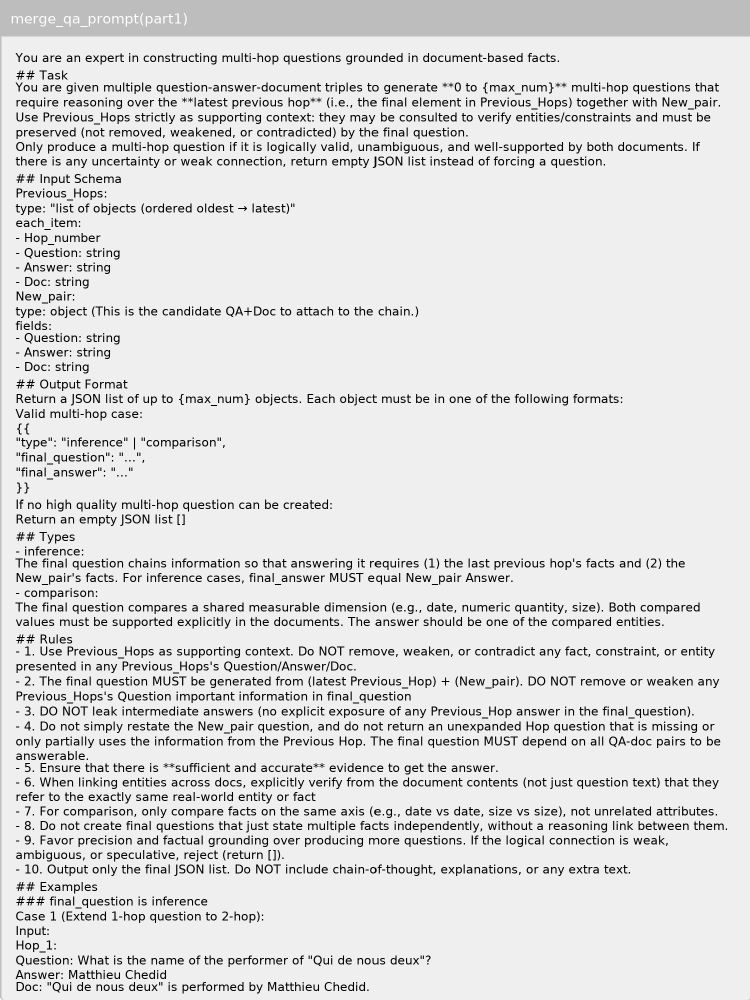}
    \caption{Merge qa prompt(part1)}
    \label{fig:merge_qa_prompt(part1)}
\end{figure*}

\begin{figure*}[!h]
    \centering
    \includegraphics[width=0.9\linewidth]{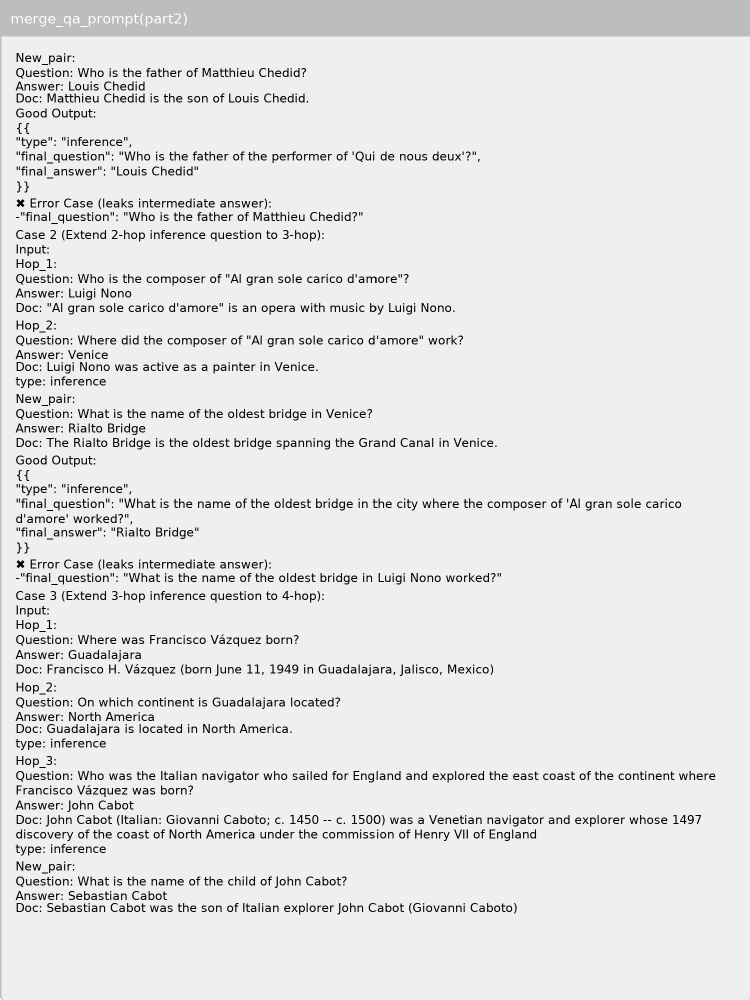}
    \caption{Merge qa prompt(part2)}
    \label{fig:merge_qa_prompt(part2)}
\end{figure*}

\begin{figure*}[!h]
    \centering
    \includegraphics[width=0.9\linewidth]{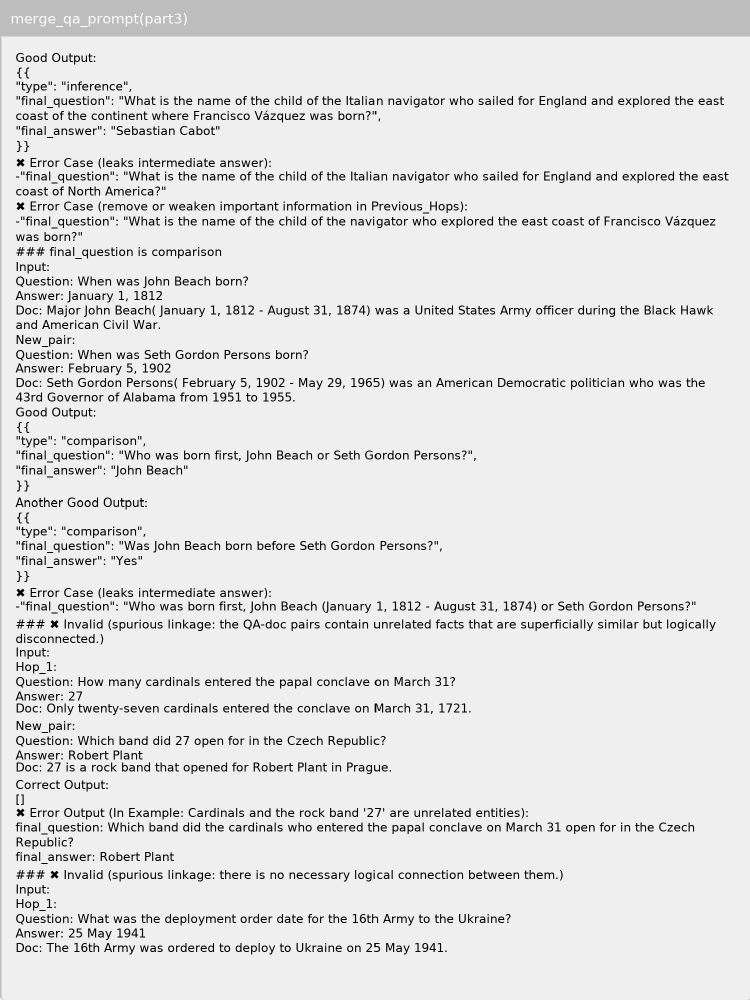}
    \caption{Merge qa prompt(part3)}
    \label{fig:merge_qa_prompt(part3)}
\end{figure*}

\begin{figure*}[!h]
    \centering
    \includegraphics[width=0.9\linewidth]{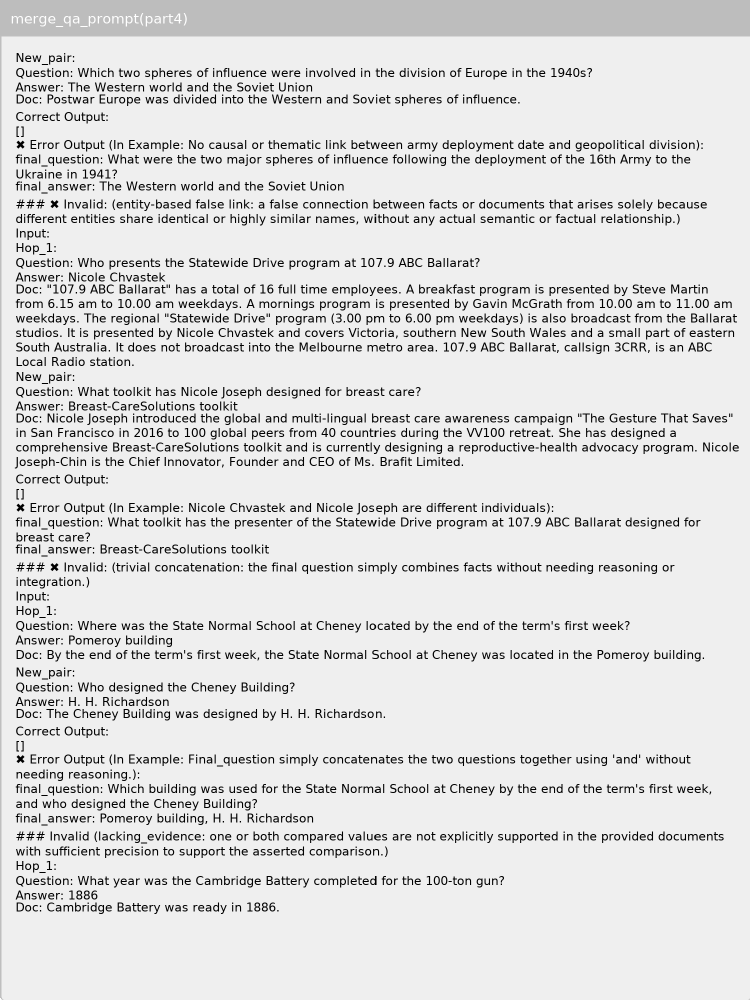}
    \caption{Merge qa prompt(part4)}
    \label{fig:merge_qa_prompt(part4)}
\end{figure*}

\begin{figure*}[!h]
    \centering
    \includegraphics[width=0.9\linewidth]{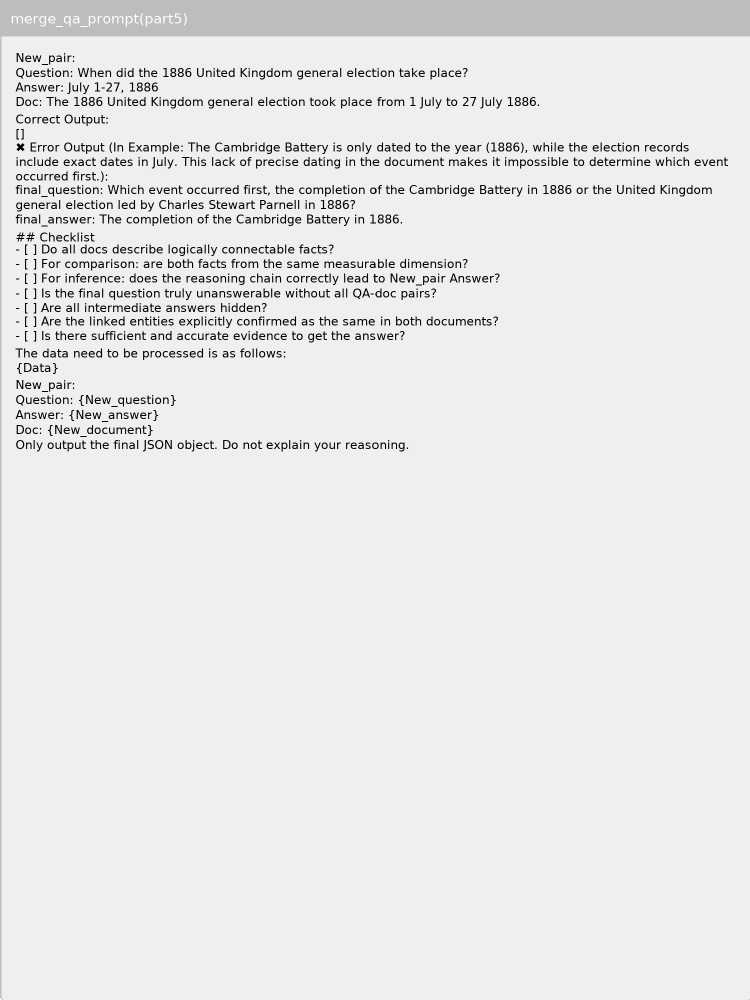}
    \caption{Merge qa prompt(part5)}
    \label{fig:merge_qa_prompt(part5)}
\end{figure*}

\begin{figure*}[!h]
    \centering
    \includegraphics[width=0.9\linewidth]{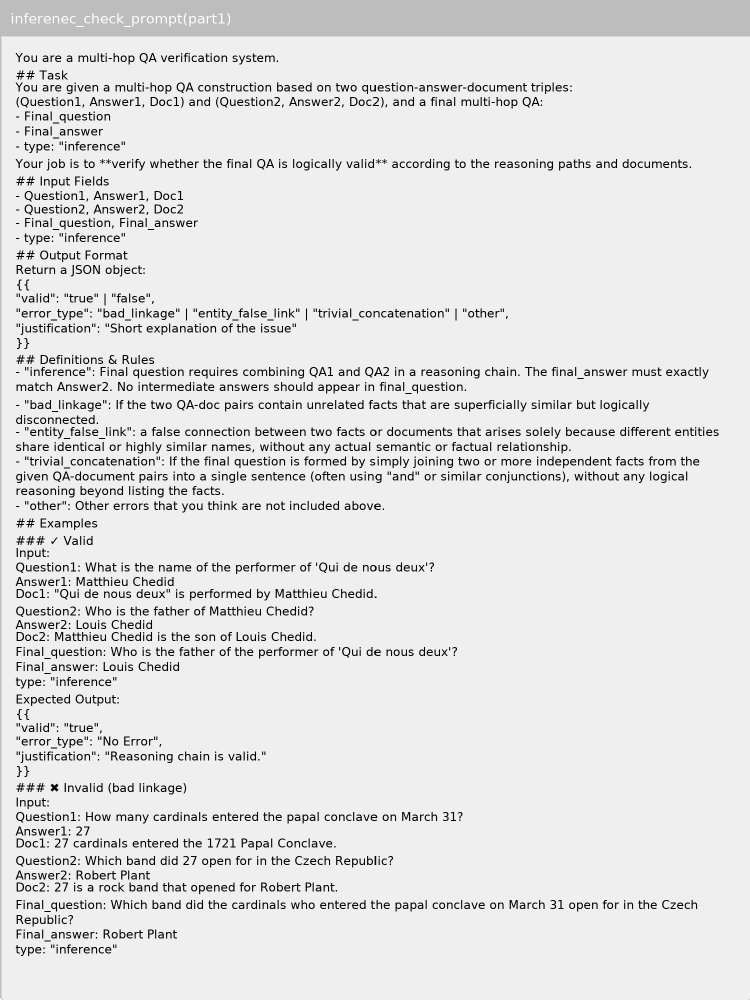}
    \caption{Inference check prompt(part1)}
    \label{fig:inference_check_prompt(part1)}
\end{figure*}

\begin{figure*}[!h]
    \centering
    \includegraphics[width=0.9\linewidth]{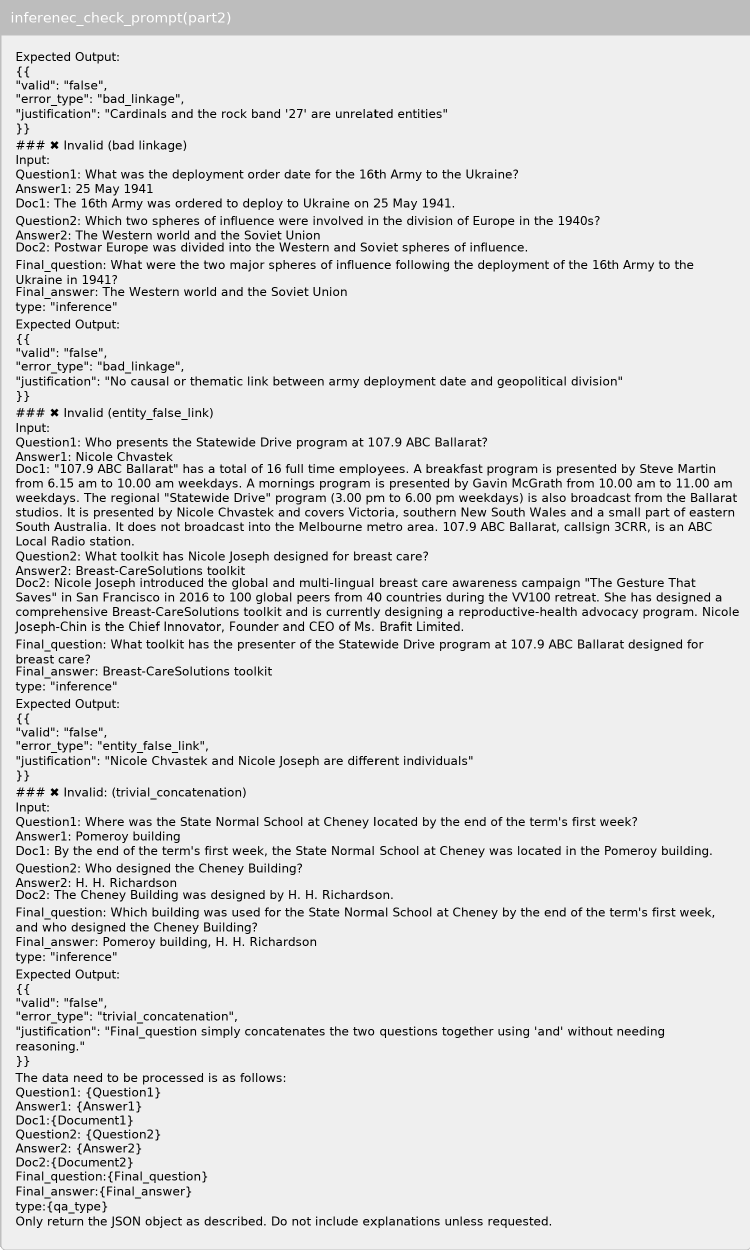}
    \caption{Inference check prompt(part2)}
    \label{fig:inference_check_prompt(part2)}
\end{figure*}

\begin{figure*}[!h]
    \centering
    \includegraphics[width=0.9\linewidth]{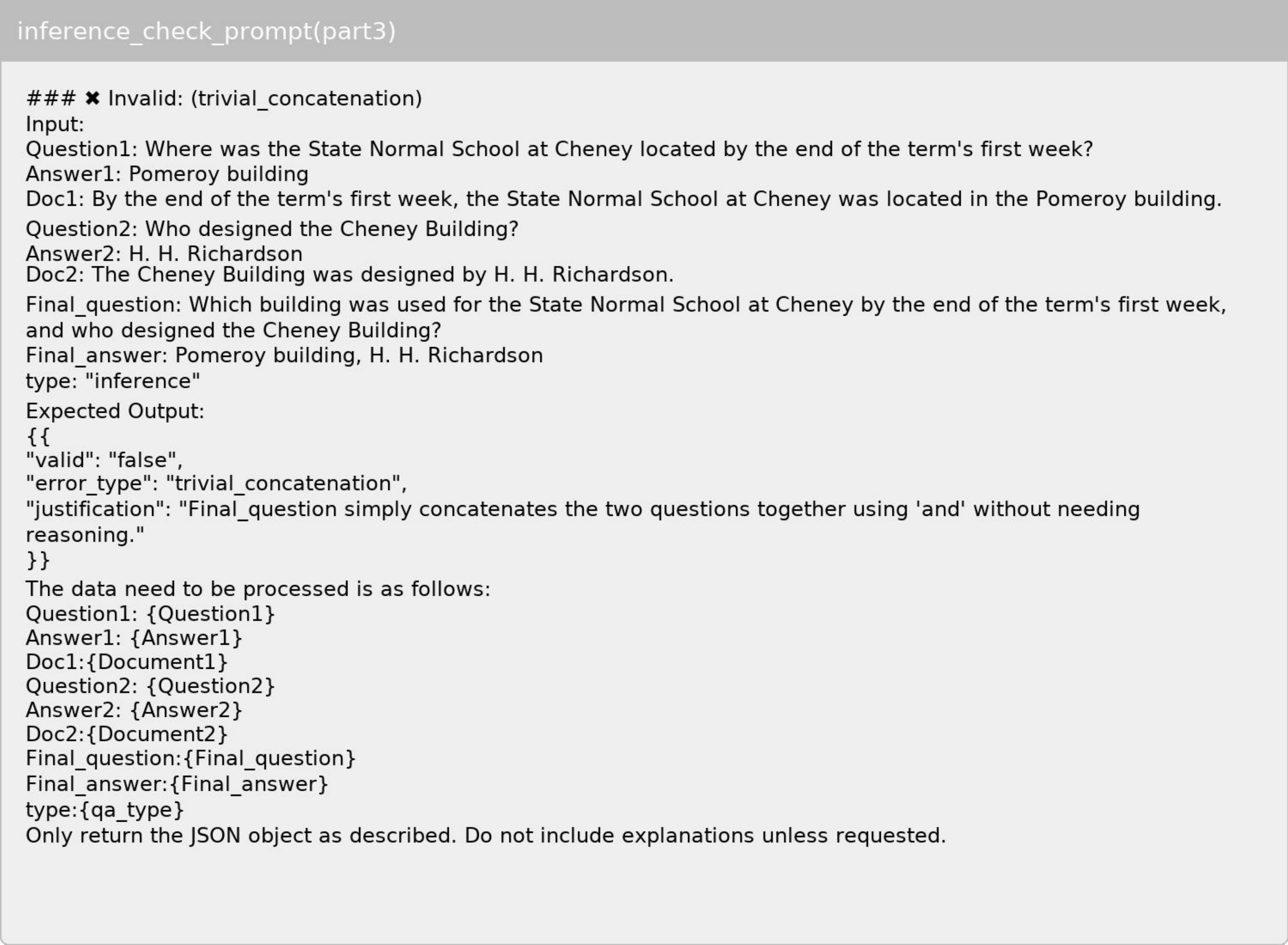}
    \caption{Inference check prompt(part3)}
    \label{fig:inference_check_prompt(part3)}
\end{figure*}

\begin{figure*}[!h]
    \centering
    \includegraphics[width=0.9\linewidth]{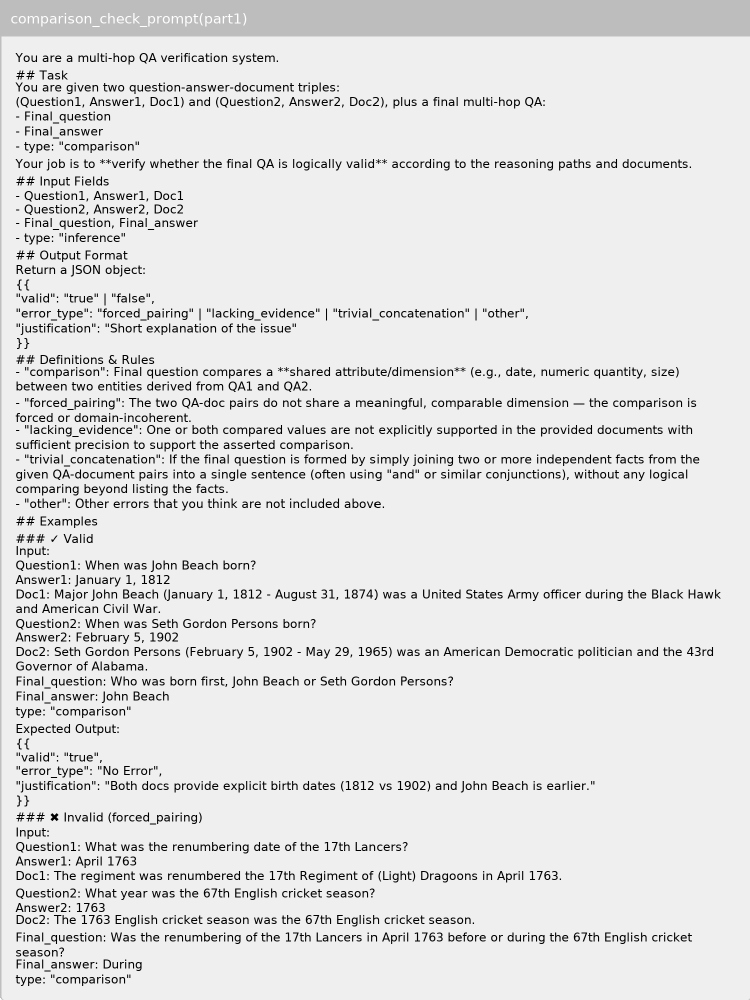}
    \caption{Comparison check prompt(part1)}
    \label{fig:comparison_check_prompt(part1)}
\end{figure*}

\begin{figure*}[!h]
    \centering
    \includegraphics[width=0.9\linewidth]{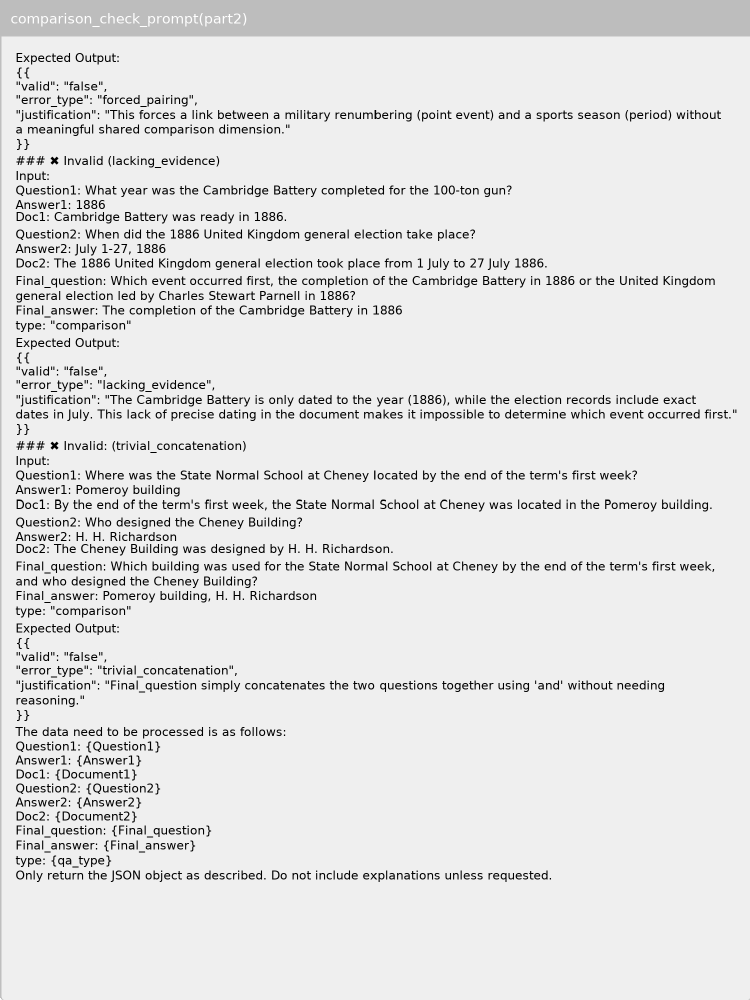}
    \caption{Comparison check prompt(part2)}
    \label{fig:comparison_check_prompt(part2)}
\end{figure*}

\begin{figure*}[!h]
    \centering
    \includegraphics[width=0.9\linewidth]{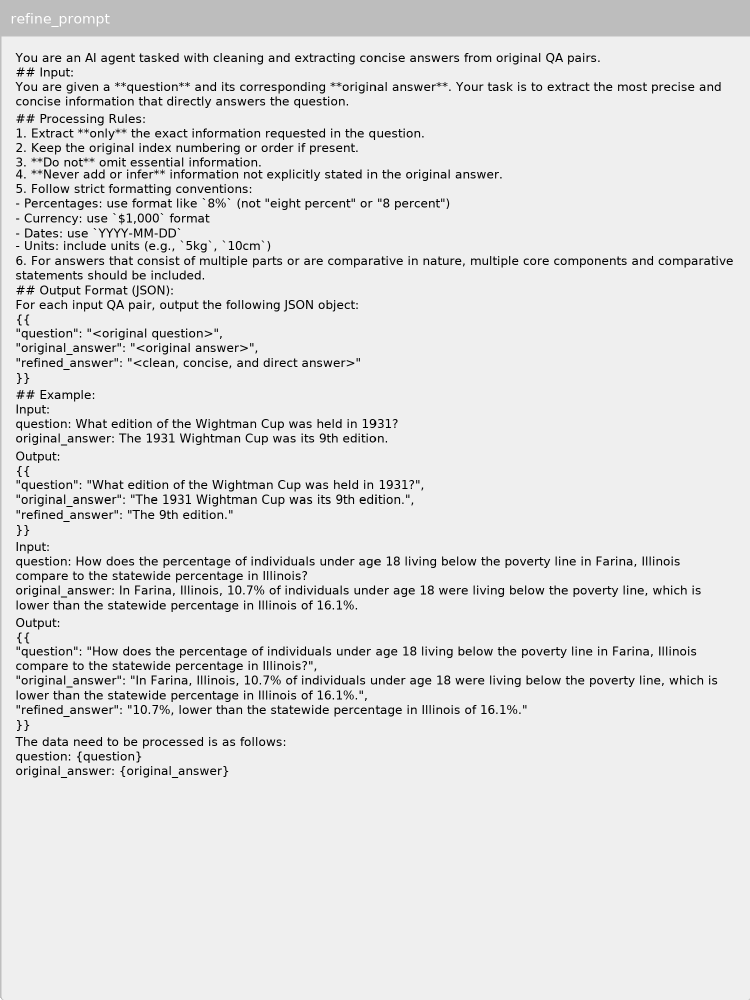}
    \caption{Refine prompt}
    \label{fig:refine_prompt}
\end{figure*}

\begin{figure*}[!h]
    \centering
    \includegraphics[width=0.9\linewidth]{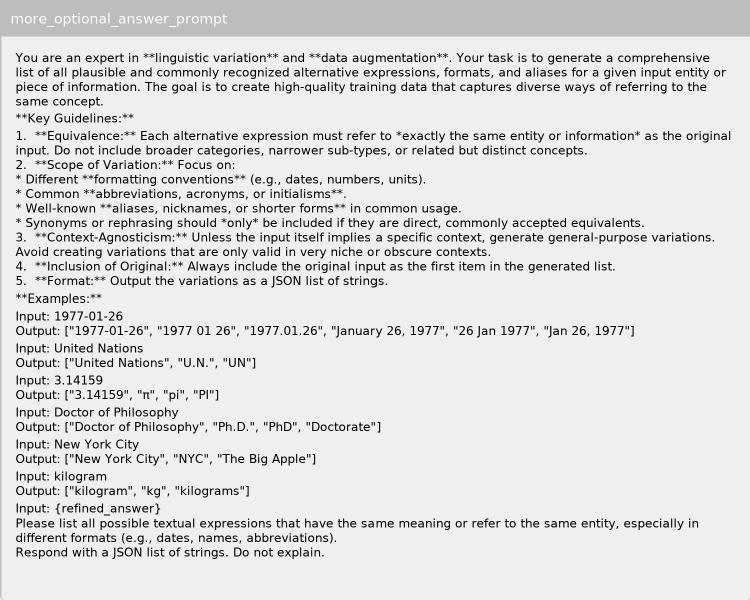}
    \caption{More optional answer prompt}
    \label{fig:more_optional_answer_prompt}
\end{figure*}

\begin{figure*}[!h]
    \centering
    \includegraphics[width=0.9\linewidth]{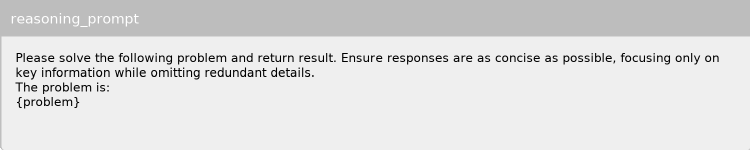}
    \caption{Reasoning prompt}
    \label{fig:reasoning_prompt}
\end{figure*}

\begin{figure*}[!h]
    \centering
    \includegraphics[width=0.9\linewidth]{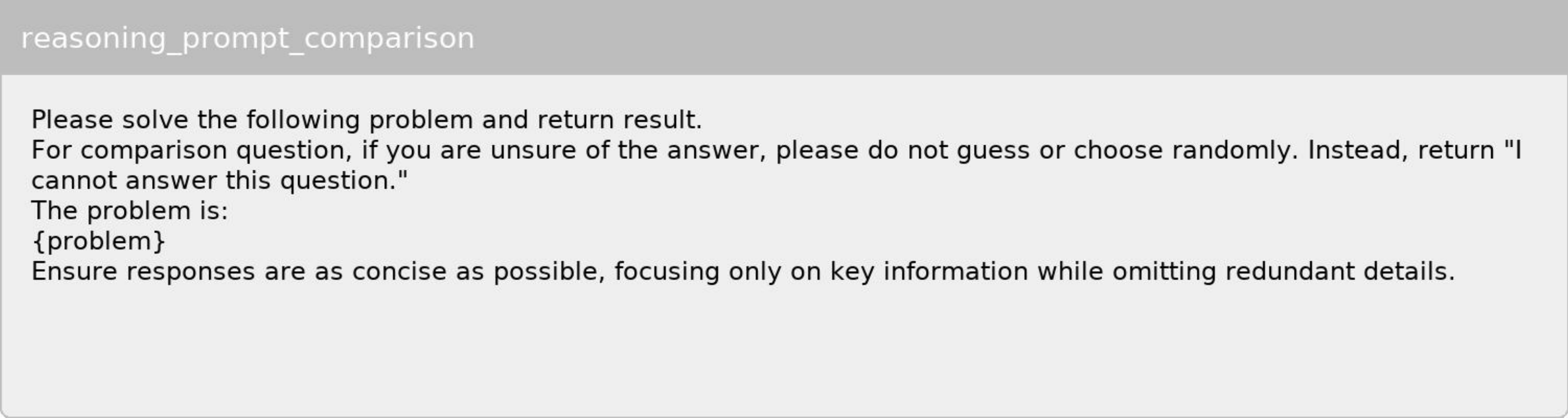}
    \caption{Reasoning prompt comparison}
    \label{fig:placeholder}
\end{figure*}

\begin{figure*}[!h]
    \centering
    \includegraphics[width=0.9\linewidth]{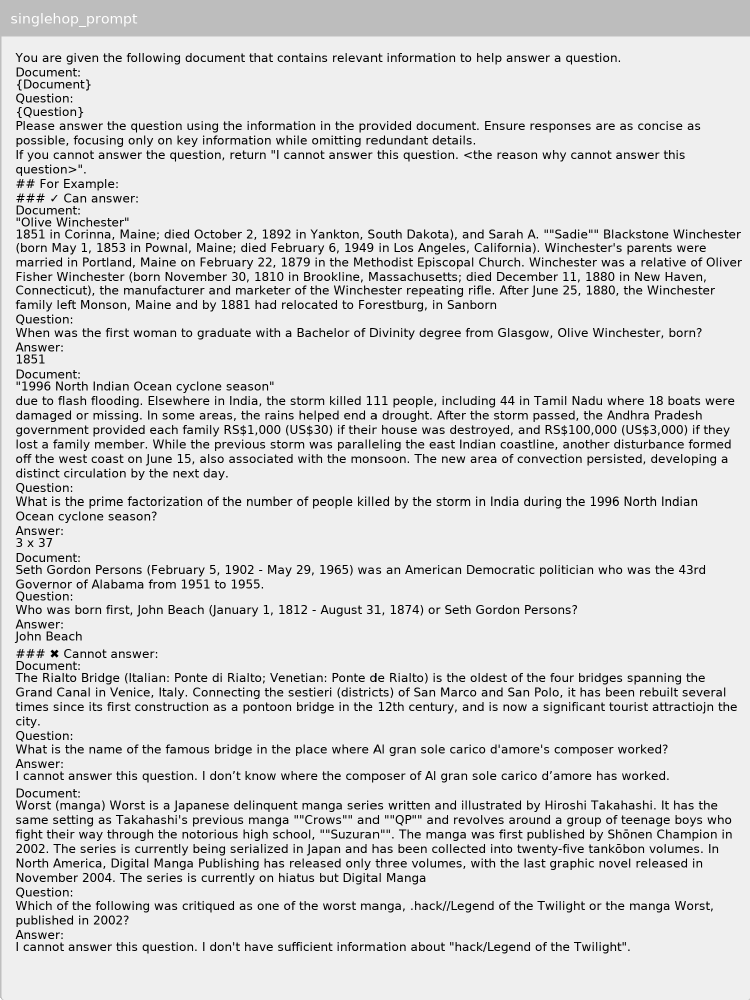}
    \caption{Singlehop prompt}
    \label{fig:singlehop_prompt}
\end{figure*}

\begin{figure*}[!h]
    \centering
    \includegraphics[width=0.9\linewidth]{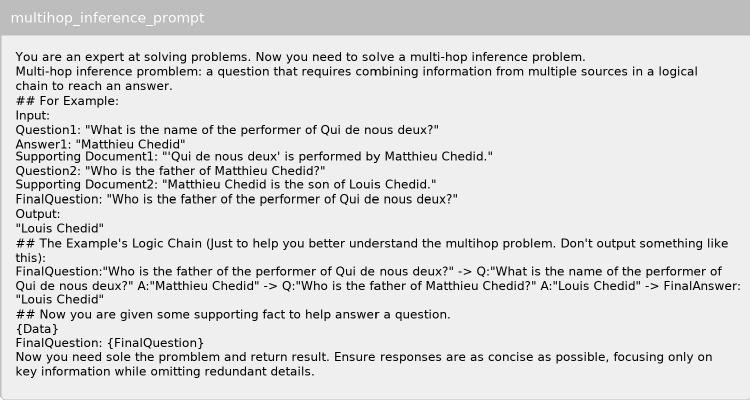}
    \caption{Multi-hop inference prompt}
    \label{fig:multihop_inference_prompt}
\end{figure*}

\begin{figure*}[!h]
    \centering
    \includegraphics[width=0.9\linewidth]{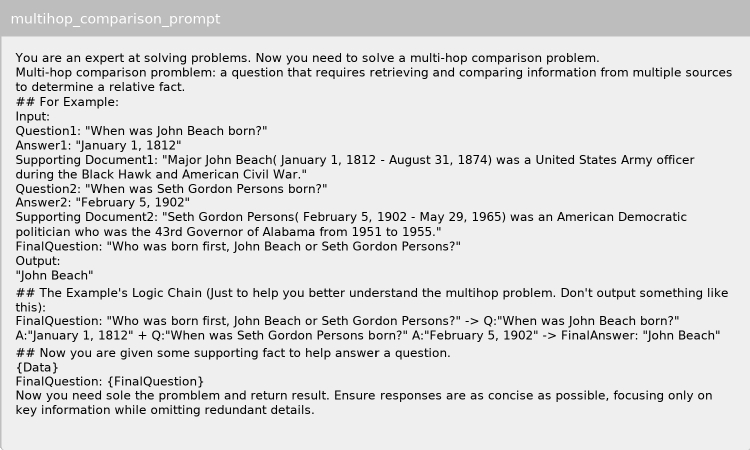}
    \caption{Multi-hop comparison prompt}
    \label{fig:multihop_comparison_prompt}
\end{figure*}

\begin{figure*}[!h]
    \centering
    \includegraphics[width=0.9\linewidth]{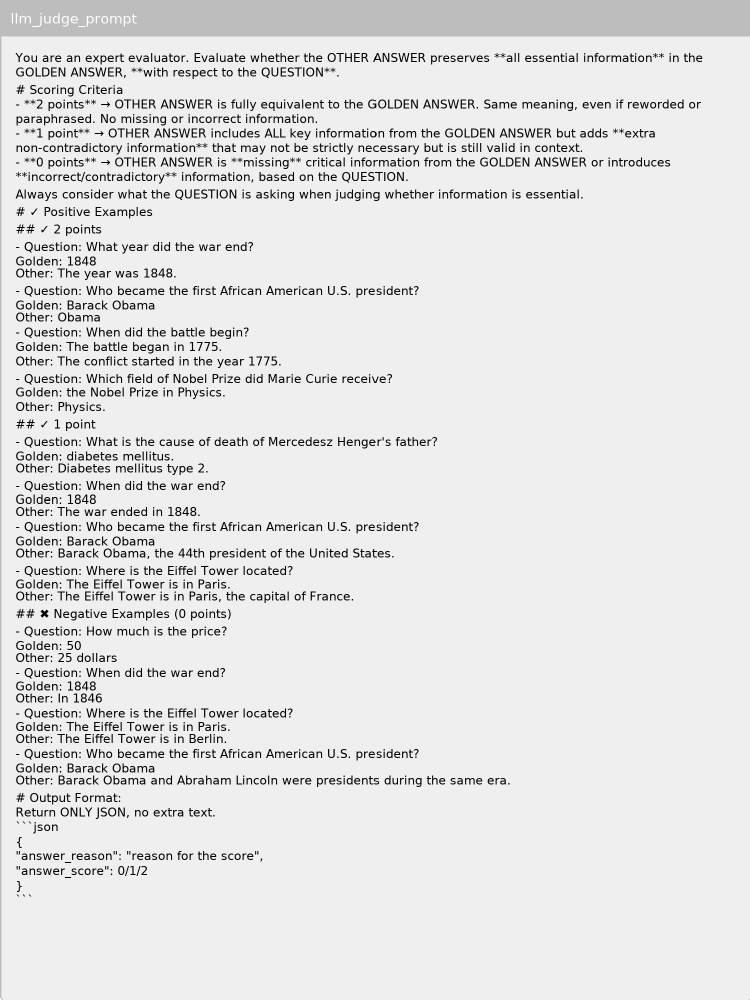}
    \caption{LLM judge prompt}
    \label{fig:llm_judge_prompt}
\end{figure*}

\begin{figure*}[!h]
    \centering
    \includegraphics[width=0.9\linewidth]{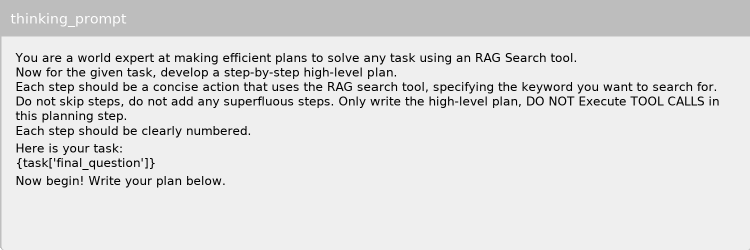}
    \caption{Thinking prompt}
    \label{fig:thinking_prompt}
\end{figure*}

\begin{figure*}[!h]
    \centering
    \includegraphics[width=0.9\linewidth]{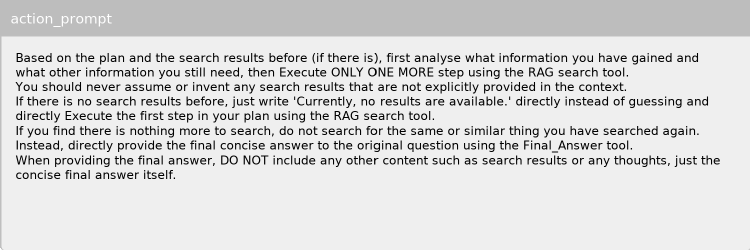}
    \caption{Action prompt}
    \label{fig:action_prompt}
\end{figure*}

\textbf{Prompts Used in Data Construction Pipeline~\ref{fig:gen_atomic_qa_prompt}--\ref{fig:llm_judge_prompt}:} \\
\textbf{Prompts Used in Evaluate~\ref{fig:thinking_prompt} and~\ref{fig:action_prompt}:} \\

\FloatBarrier

\FloatBarrier
The LLM judge prompt is the same as the prompt used in the data construction process.


\section{Error Case Analysis}
\label{app:error_case}
To ensure conciseness and highlight the key points, we present below specific model output examples for the \textbf{Error Case Study}, with some non-essential parts omitted.

\begin{table}[!htbp] 
\centering
\renewcommand{\arraystretch}{1.1} 
\small
\begin{tabularx}{\linewidth}{>{\raggedright\arraybackslash}p{\linewidth}}
\toprule
\textbf{User:} \\
You are a world expert at making efficient plans to solve any task using an RAG Search tool. ... \\
Here is your task: Which broader trading area does the metropolitan area of the city where the professional football team that faced the player who kicked a 27-yard field goal early in the final period for the 2009 UCLA Bruins football team in his sixth career game-winner is based form the core of? \\
\midrule

\textbf{Assistant:} \\
1. Search RAG for “2009 UCLA Bruins 27-yard field goal early in the final period sixth career game-winner” to identify the player and the specific game.  \\
2. Search RAG for “[Player Name] first NFL game opponent” to determine which professional football team he first faced.  \\
3. Search RAG for “[Opponent Team Name] home city” to find the city where that professional team is based.  \\
4. Search RAG for “[City Name] metropolitan area” to get the official metropolitan area designation.  \\
5. Search RAG for “[Metropolitan Area Name] forms the core of” to identify the broader trading area that metro area forms the core of. \\
\midrule

\textbf{User:} \\
Based on the plan and the search results before (if there is), first analyse what information you have gained and what other information you still need, then EXECUTE ONLY ONE MORE step using the RAG search tool. ... \\
\midrule

\textbf{Assistant:} \\
\texttt{"function\_call": \{"name": "RAG\_search", "arguments": "\{query: 2009 UCLA Bruins 27-yard field goal early in the final period sixth career game-winner, topk: 5\}"\}} \\
\midrule

\textbf{[RAGTool Results]} \\
Query: "2009 UCLA Bruins 27-yard field goal early in the final period sixth career game-winner" \\
Topk = 5 \\
Document: \\
"2009 UCLA Bruins football team" ...
 Kai Forbath kicked a 27-yard field goal early in the final period \\
\midrule

\textbf{User:} \\
Based on the plan and the search results before (if there is), first analyse what information you have gained and what other information you still need, then EXECUTE ONLY ONE MORE step using the RAG search tool. ... \\
\midrule

\textbf{Assistant:} \\
\texttt{"function\_call": \{"name": "RAG\_search", "arguments": "\{query: Kai Forbath first NFL game opponent, topk: 5\}"\}} \\
\midrule

\textbf{[RAGTool Results]} \\
Query: "Kai Forbath first NFL game opponent" \\
Topk = 5 \\
Document: \\
Kai Forbath" ... Forbath made his NFL and Redskins debut knocking through a 50-yard attempt as his first career NFL field goal in the home game against the Minnesota Vikings in Week 6 on October 14, 2012. \\
\midrule

\textbf{ ... As a consequence, the entire subsequent reasoning trajectory deviates accordingly.} \\
\midrule

\textbf{Assistant: }\\
Final\_Answer: Minneapolis–St. Paul MN–WI Combined Statistical Area \\

\bottomrule
\end{tabularx}
\caption{Error Case}
\label{tab:error_case}
\end{table}

\begin{table}[!htbp]
\centering
\renewcommand{\arraystretch}{1.1}
\small
\begin{tabularx}{\linewidth}{>{\raggedright\arraybackslash}p{\linewidth}}
\toprule
\textbf{Hop1:} \\
\textbf{Atomic Question:} Who kicked a 27-yard field goal early in the final period for the 2009 UCLA Bruins football team? \\
\textbf{Atomic Answer: } Kai Forbath \\
\textbf{Doc:} "2009 UCLA Bruins football team" ...
 Kai Forbath kicked a 27-yard field goal early in the final period \\
\midrule

\textbf{Hop2:} \\
\textbf{Atomic Question:} Against which team did Kai Forbath achieve his sixth career game-winner? \\
\textbf{Atomic Answer: } Atlanta Falcons \\
\textbf{Doc:} "Kai Forbath" ... season finale in Week 17 against the Atlanta Falcons as time expired, his sixth career game-winner. \\
\textbf{Final Question:} Against which team did the player who kicked a 27-yard field goal early in the final period for the 2009 UCLA Bruins football team achieve his sixth career game-winner? \\
\textbf{Final Answer:} Atlanta Falcons \\
\midrule

\textbf{Hop3:} \\
\textbf{Atomic Question:} What is the location of the Atlanta Falcons' professional football team? \\
\textbf{Atomic Answer: } Atlanta, Georgia \\
\textbf{Doc:} "Atlanta Falcons" ... Atlanta Falcons The Atlanta Falcons are a professional American football team based in Atlanta, Georgia.\\
\textbf{Final Question:} What is the location of the professional football team that faced the player who kicked a 27-yard field goal early in the final period for the 2009 UCLA Bruins football team in his sixth career game-winner? \\
\textbf{Final Answer:} Atlanta, Georgia \\
\midrule

\textbf{Hop4:} \\
\textbf{Atomic Question:} Which broader trading area does the Atlanta, Georgia metropolitan area form the core of? \\
\textbf{Atomic Answer: } Atlanta–Athens-Clarke–Sandy Springs Combined Statistical Area \\
\textbf{Doc:} "Atlanta metropolitan area" ... The metro area forms the core of a broader trading area, the Atlanta–Athens-Clarke–Sandy Springs Combined Statistical Area. The Combined Statistical Area spans up to 39 counties in north Georgia and has an \\
\textbf{Final Question:} Which broader trading area does the metropolitan area of the city where the professional football team that faced the player who kicked a 27-yard field goal early in the final period for the 2009 UCLA Bruins football team in his sixth career game-winner is based form the core of? \\
\textbf{Final Answer:} Atlanta–Athens-Clarke–Sandy Springs Combined Statistical Area \\

\bottomrule
\end{tabularx}
\caption{Golden Trace}
\label{tab:golden_trace}
\end{table}

For Table ~\ref{tab:error_case}~\ref{tab:golden_trace}, it can be observed that the model incorrectly transformed \emph{``sixth career game-winner"} in the second step into \emph{``the professional football team he first faced"}, thereby causing the entire reasoning trajectory to deviate.

\FloatBarrier
\section{Human Evaluation Details and Ethical Considerations}
\label{app:human_eval}
\textbf{Annotator Recruitment.} 
Human evaluators were recruited from adult participants fluent in English. All annotators were informed of the study's purpose and participated voluntarily. To ensure the reliability of judgments and high-quality annotations, we conducted a comprehensive training session for all participants to ensure they fully understood the guidelines and could correctly perform the evaluation tasks.

\textbf{Annotation Procedure and Criteria.} 
Annotators were presented with data samples requiring human verification. The task was designed as a binary classification problem: annotators were asked to either ``Retain'' or ``Discard'' the sample based on strict quality standards. The evaluation criteria consisted of two main categories: \textit{General Quality Checks:} Annotators verified Factuality \& Faithfulness by ensuring the answer was strictly grounded in the source text and free from any information fabrication or hallucination. They also checked Logic \& Fluency to ensure questions were grammatically natural and logically sound. \textit{Type-Specific Checks:} For Reasoning Questions, annotators reviewed the reasoning chain layer by layer to ensure that every intermediate step constituted a valid deduction. For Comparison Questions, they verified the consistency of the comparison dimensions and the factual accuracy of the final conclusion.

\textbf{Inter-Annotator Agreement and Adjudication.} 
All data samples requiring human verification were independently evaluated by three annotators. To assess the reliability of the annotation process, we calculated the Inter-Annotator Agreement (IAA) using Fleiss' Kappa ($\kappa$). The resulting $\kappa$ score was 0.65, indicating substantial agreement among annotators. For cases where annotators disagreed (i.e., non-unanimous decisions), the final labels were determined through a consensus discussion involving both the annotators and the authors to ensure the highest data quality.

\textbf{Compensation.} 
Annotators were compensated at a fixed rate of \$15 per hour, which is consistent with or above standard minimum wage guidelines. The compensation was independent of the label distribution to avoid incentive bias.

\textbf{Ethical Considerations.} 
The annotation process did not involve the collection of any personally identifiable information (PII). All content was anonymized. Given the non-invasive nature of the task, this study aligns with standard ethical guidelines for NLP data annotation.

\end{document}